\documentclass{article}

\usepackage{graphicx}
\usepackage{subfig}

\usepackage{booktabs}
\usepackage{multirow}
\usepackage{rotating}

\usepackage{color}

\usepackage[T1]{fontenc}
\usepackage{amsmath}
\usepackage{amssymb}
\usepackage{nicefrac}
\usepackage{microtype}

\DeclareMathOperator{\cov}{cov}
\newcommand{\bb}[1]{\mathbb{#1}}
\newcommand{\R}{\bb{R}}
\newcommand{\cm}[1]{\mathcal{#1}}
\newcommand{\bm}[1]{\mathbf{#1}}

\renewcommand{\epsilon}{\varepsilon}
\newcommand{\acro}[1]{\textsc{\MakeLowercase{#1}}}
\newcommand{\trans}{\ensuremath{^\top}}
\DeclareMathOperator{\diag}{diag}

\pdfoutput=1
\usepackage{hyperref}
\hypersetup{
  colorlinks=true,
  linkcolor=black,
  citecolor=black,
  filecolor=black,
  urlcolor=black
}

\usepackage{enumitem}
\usepackage{url}
\usepackage{lscape}

\newtheorem{my_def}{Definition}
\newtheorem{lem}{Lemma}

\usepackage{algorithm}
\usepackage{algpseudocode}

\usepackage[osf]{libertine}
\usepackage[a4paper, margin=1.35in]{geometry}
\usepackage[usenames,dvipsnames]{xcolor}
\usepackage{color}
\usepackage{tikz}
\usepackage{verbatim}
\usetikzlibrary{shapes,arrows,positioning,backgrounds,fit}
\pgfdeclarelayer{bg}
\pgfsetlayers{bg,main}

\usepackage{usbib}
\begin{document}

\title{Propagation Kernels}

\author{Marion Neumann      \and
        Roman Garnett       \and
        Christian Bauckhage \and
        Kristian Kersting
}

\maketitle

\begin{abstract}
  We introduce \emph{propagation kernels,} a general graph-kernel
  framework for efficiently measuring the similarity of structured
  data. Propagation kernels are based on monitoring how information
  spreads through a set of given graphs.  They leverage early-stage
  distributions from propagation schemes such as random walks to capture
  structural information encoded in node labels, attributes, and edge
  information. This has two benefits. First, off-the-shelf propagation
  schemes can be used to naturally construct kernels for many graph
  types, including labeled, partially labeled, unlabeled, directed, and
  attributed graphs. Second, by leveraging existing efficient and
  informative propagation schemes, propagation kernels can be
  considerably faster than state-of-the-art approaches without
  sacrificing predictive performance.  We will also show that if the
  graphs at hand have a regular structure, for instance when modeling
  image or video data, one can exploit this regularity to scale the
  kernel computation to large databases of graphs with thousands of
  nodes.  We support our contributions by exhaustive experiments on a
  number of real-world graphs from a variety of application domains.
\end{abstract}

\section{Introduction}
\label{sec:intro}
Learning from structured data is an active area of research.  As
domains of interest become increasingly diverse and complex, it is
important to design flexible and powerful methods for analysis and
learning.  By \emph{structured data} we refer to situations where
objects of interest are structured and hence can naturally be
represented using graphs.  Real-world examples are molecules or
proteins, image annotated with semantic information, text documents
reflecting complex content dependencies, and manifold data modeling
objects and scenes in robotics.  The goal of learning with graphs is
to exploit the rich information contained in graphs representing
structured data. The main challenge is to efficiently exploit the
graph structure for machine-learning tasks such as classification or
retrieval.  A popular approach to learning from structured data is to
design graph kernels measuring the similarity between graphs.  For
classification or regression problems, the graph kernel can then be
plugged into a kernel machine, such as a support vector machine or a
Gaussian process, for efficient learning and prediction.

Several graph kernels have been proposed in the literature, but they
often make strong assumptions as to the nature and availability of
information related to the graphs at hand. The most simple of these
proposals assume that graphs are unlabeled and have no structure
beyond that encoded by their edges.  However, graphs encountered in
real-world applications often come with rich additional information
attached to their nodes and edges.  This naturally implies many
challenges for representation and learning such as:
\begin{itemize}[noitemsep]
 \item missing information leading to partially labeled graphs,
 \item uncertain information arising from aggregating information from multiple sources, and
 \item continuous information derived from complex and possibly noisy sensor measurements.
\end{itemize}
Images, for instance, often have metadata and semantic annotations
which are likely to be only partially available due to the high cost
of collecting training data. Point clouds captured by laser range
sensors consist of continuous 3\textsc{d} coordinates and curvature
information; in addition, part detectors can provide possibly noisy
semantic annotations.  Entities in text documents can be backed by
entire Wikipedia articles providing huge amounts of structured
information, themselves forming another network.

Surprisingly, existing work on graph kernels does not broadly account
for these challenges.  Most of the existing graph
kernels~\citep{shervashidzeVPMB09,ShervashidzeB11,HidoK09,KashimaTI03,GartnerFW03}
are designed for unlabeled graphs or graphs with a complete set of
discrete node labels.  Kernels for graphs with continuous node
attributes have only recently gained greater interest
\citep{borgwardtK05,KriegeM12,FeragenKPBB13}.  These graph kernels have
two major drawbacks: they can only handle graphs with complete label
or attribute information in a principled manner and they are either
efficient, but limited to specific graph types, or they are flexible,
but their computation is memory and/or time consuming.  To overcome
these problems, we introduce \emph{propagation kernels}.  Their design
is motivated by the observation that iterative information propagation
schemes originally developed for within-network relational and
semi-supervised learning have two desirable properties: they capture
structural information and they can often adapt to the aforementioned
issues of real-world data. In particular, propagation schemes such as
diffusion or label propagation can be computed efficiently and they
can be initialized with uncertain and partial information.

A high-level overview of the propagation kernel algorithm is as
follows.  We begin by initializing label and/or attribute
distributions for every node in the graphs at hand.  We then
iteratively propagate this information along edges using an
appropriate propagation scheme.  By maintaining entire distributions
of labels and attributes, we can accommodate uncertain information in
a natural way.  After each iteration, we compare the similarity of the
induced node distributions between each pair of graphs.  Structural
similarities between graphs will tend to induce similar local
distributions during the propagation process, and our kernel will be
based on approximate counts of the number of induced similar
distributions throughout the information propagation.

To achieve competitive running times and to avoid having to compare
the distributions of all pairs of nodes between two given graphs, we
will exploit \emph{locality sensitive hashing} (\acro{lsh}) to bin the
label/attribute distributions into efficiently computable graph
feature vectors in time linear in the total number of nodes.  These
new graph features will then be fed into a base kernel, a common
scheme for constructing graph kernels. Whereas \textsc{lsh} is usually
used to preserve the $\ell^1$ or $\ell^2$ distance, we are able to
show that the hash values can preserve both the total variation and
the Hellinger probability metrics.  Exploiting explicit feature
computation and efficient information propagation, propagation kernels
allow for using graph kernels to tackle novel applications beyond the
classical benchmark problems on datasets of chemical compounds and
small- to medium-sized image or point-cloud graphs.

Propagation kernels were originally defined and applied for graphs
with discrete node labels \citep{NeumannPGK12,Neumann13mlg}; here we
extend their definition to a more general and flexible framework that
is able to handle continuous node attributes.  In addition to this
expanded view of propagation kernels, we also introduce and discuss
efficient propagation schemes for numerous classes of graphs.  A
central message is: \vspace{0.1cm} \\
\centerline{\emph{A suitable propagation scheme is the key to designing}}
\vspace{0.1cm}
\centerline{\emph{fast and powerful propagation kernels.}}

In particular, we will discuss propagation schemes applicable to huge
graphs with regular structure, for example grid graphs representing
images or videos. Thus, implemented with care, propagation kernels can
easily scale to large image databases.  The design of kernels for
grids allows us to perform graph-based image analysis not only on the
scene level \citep{NeumannPGK12,HarchaouiB07} but also on the pixel
level opening up novel application domain for graph kernels.

We proceed as follows. We begin by touching upon related work on
kernels and graphs. After introducing information propagation on
graphs via random walks, we introduce the family of propagation
kernels (Section~\ref{sec:PKs}).  The following two sections discuss
specific examples of the two main components of propagation kernels:
node kernels for comparing propagated information
(Section~\ref{sec:node_kernel}) and propagation schemes for various
kinds of information (Section~\ref{sec:Propagation_Schemes}).  We will
then analyze the sensitivity of propagation kernels with respect to
noisy and missing information, as well as with respect to the choice
of their parameters.  Finally, to demonstrate the feasibility and
power of propagation kernels for large real-world graph databases, we
provide experimental results on several challenging classification
problems, including commonly used bioinformatics benchmark problems,
as well as real-world applications such as image-based plant-disease
classification and 3\textsc{d} object category prediction in the
context of robotic grasping.

\section{Kernels and Graphs}
\label{sec:related_work}
Propagation kernels are related to three lines of research on kernels:
kernels between graphs (\emph{graph kernels}) developed within the
graph mining community, kernels between nodes (\emph{kernels on
  graphs}) established in the within-network relational learning and
semi-supervised learning communities, and kernels between probability
distributions.

\subsection{Graph Kernels}
Propagation kernels are deeply connected to several graph kernels
developed within the graph-mining community.  Categorizing graph
kernels with respect to how the graph structure is captured, we can
distinguish four classes: kernels based on
walks~\citep{GartnerFW03,KashimaTI03,vishwanathanSKB10,HarchaouiB07}
and paths~\citep{borgwardtK05,FeragenKPBB13}, kernels based on
limited-size
subgraphs~\citep{horvathGW04,shervashidzeVPMB09,KriegeM12}, kernels
based on subtree patterns~\citep{maheV09,Ramon03}, and kernels based on
structure propagation~\citep{ShervashidzeB11}.  However, there are two
major problems with most existing graph kernels: they are often slow
or overly specialized.  There are efficient graph kernels specifically
designed for unlabeled and fully labeled
graphs~\citep{shervashidzeVPMB09,ShervashidzeB11}, attributed
graphs~\citep{FeragenKPBB13}, or planar labeled
graphs~\citep{HarchaouiB07}, but these are constrained by design.
There are also more flexible but slower graph kernels such as the
shortest path kernel~\citep{borgwardtK05} or the common subgraph
matching kernel~\citep{KriegeM12}.

The Weisfeiler--Lehman (\textsc{wl}) subtree kernel, one instance of
the recently introduced family of
\textsc{wl}-kernels~\citep{ShervashidzeB11}, computes count features
for each graph based on signatures arising from iterative multi-set
label determination and compression steps. In each kernel iteration,
these features are then fed into a base kernel.  The
\textsc{wl}-kernel is finally the sum of those base kernels over the
iterations.

Although \acro{wl}-kernels are usually competitive in terms of
performance and runtime, they are designed for fully labeled graphs.
The challenge of comparing large, \emph{partially} labeled graphs --
as considered by propagation kernels introduced in the present paper
-- remains to a large extent unsolved. One proposal for extending a
graph kernel to the partially labeled case is to mark unlabeled nodes
with a unique symbol, as suggested in \citep{ShervashidzeB11}.
However, collapsing all unlabeled nodes into a single label neglects
any notion of uncertainty in the labels.  Another option is to
propagate labels across the graph and then run a graph kernel on the
imputed labels \citep{NeumannPGK12}. Unfortunately, this also ignores
the uncertainty induced by the inference procedure, as hard labels
have to be assigned after convergence.  A key observation motivating
propagation kernels is that intermediate label distributions induced
well before convergence carry information about the structure of the
graph.  Propagation kernels interleave label inference and kernel
computation steps, avoiding the requirement of running inference to
termination.

\subsection{Kernels on Graphs and Within-network Relational Learning}
Measuring the structural similarity of local node neighborhoods has
recently become popular for inference in networked
data~\citep{KondorL02,Desrosiers09,NeumannGK13} where this idea has
been used for designing kernels on graphs (kernels between the nodes
of a graph) and for within-network relational learning approaches.  An
example of the former are \emph{coinciding walk
  kernels}~\citep{NeumannGK13} which are defined in terms of the
probability that the labels encountered during parallel random walks
starting from the respective nodes of a graph coincide.  Desrosiers et
al. \citep{Desrosiers09} use a similarity measure based on parallel
random walks with constant termination probability in a
relaxation-labeling algorithm.  Another approach exploiting random
walks and the structure of subnetworks for node-label prediction is
\emph{heterogeneous label propagation} \citep{HwangK10}.  \emph{Random
  walks with restart} are used as proximity weights for so-called
``ghost edges'' in \citep{GallagherGhost08}, but then the features
considered by a later bag of logistic regression classifiers are only
based on a one-step neighborhood.  The connection between these
approaches and propagation kernels, which is based on the use of
random walks to measure structure similarity, constitutes an important
contact point of graph-based machine learning for inference about
node- and graph-level properties.

\subsection{Kernels between Probability Distributions and Kernels between Sets}
Finally, propagation kernels mark another contact point, namely
between graph kernels and kernels between probability
distributions~\citep{JaakkolaH98,laffertyL02,morenoHV03,jebaraKH04} and
between sets~\citep{KondorJ03,ShiPDLSV09}.  However, whereas the former
essentially build kernels based on the outcome of probabilistic
inference after convergence, propagation kernels intuitively count
common sub-distributions induced after each iteration of running
inference in two graphs.

Kernels between sets and more specifically between structured sets,
also called \emph{hash kernels}~\citep{ShiPDLSV09}, have been
successfully applied to strings, data streams, and unlabeled
graphs. While propagation kernels hash probability distributions and
derive count features from them, hash kernels directly approximate the
kernel values $k(x,x')$, where $x$ and $x'$ are (structured) sets.
Propagation kernels iteratively approximate node kernels $k(v_i,v_j)$
comparing a node $v_i$ in graph $G^{(i)}$ with a node $v_j$ in graph
$G^{(j)}$.  Counts summarizing these approximations are then fed into
a base kernel that is computed exactly.  Before we give a detailed
definition of propagation kernels, we introduce the basic concept of
information propagation on graphs, and exemplify important propagation
schemes and concepts when utilizing random walks for learning with
graphs.

\section{Information Propagation on Graphs}
\label{sec:RWs}
Information propagation or diffusion on a graph is most commonly
modeled via Markov random walks (\textsc{rw}s).  Propagation kernels
measure the similarity of two graphs by comparing node label or
attribute distributions after each step of an appropriate random
walk. In the following, we review label diffusion and label
propagation via \textsc{rw}s -- two techniques commonly used for
learning on the node level \citep{ZhuGL03,SzummerJ01}.  Based on these
ideas, we will then develop propagation kernels in the subsequent
sections.

\subsection{Basic Notation}
Throughout, we consider graphs whose nodes are endowed with (possibly
partially observed) label and/or attribute information. That is, a
graph $G=(V,E,\ell)$ is represented by a set of $|V|=n$ vertices, a
set of edges $E$ specified by a weighted adjacency matrix $A \in
\bb{R}^{n \times n}$, and a label function $\ell\colon V \rightarrow
\cm{L}$ with $\cm{L} = \left( [k], \R^D \right)$, where $k$ is the
number of available node labels and $D$ is the dimension of the
continuous attributes.  Given $V = \{v_1,v_2,...,v_n\}$, \emph{node
  labels} $\ell(v_i)$ are represented by nominal values and
\emph{attributes} $\mathbf{x}_i \in \bb R^D$ are represented by
continuous vectors.

\subsection{Markov Random Walks}
Consider a graph $G = (V,E)$. A \emph{random walk} on $G$ is a Markov
process $X = \{X_t : t\geq0 \}$ with a given initial state $X_0 =
v_i$.  We will also write $X_t^{(i)}$ to indicate that the walk began
at $v_i$.  The transition probability $T_{ij} = P(X_{t+1}=v_j \mid
X_{t}=v_i)$ only depends on the current state $X_t=v_i$ and the
one-step transition probabilities for all nodes in $V$ can be easily
represented by the row-normalized adjacency or \emph{transition
  matrix} $T = D^{-1}A$, where $D = \diag(\sum_j A_{ij})$.

\subsection{Information Propagation via Random Walks}
For now, we consider (partially) labeled graphs without attributes,
where $V = V_L \cup V_U$ is the union of labeled and unlabeled nodes
and $\ell\colon V \rightarrow [k]$ is a label function with known
values for the nodes in $V_L$.  A common mechanism for providing
labels for the nodes of an unlabeled graph is to define the label
function by $\ell(v_i) = \sum_j A_{ij} = \text{degree}(v_i)$.  Hence
for fully labeled and unlabeled graphs we have $V_U = \emptyset$.  We
will monitor the distribution of labels encountered by random walks
leaving each node in the graph to endow each node with a
$k$-dimensional feature vector.  Let the matrix $P_0 \in \bb R ^{n
  \times k}$ give the prior label distributions of all nodes in $V.$
If node $v_i \in V_L$ is observed with label $\ell(v_i)$, then the
$i$th row in $P_0$ can be conveniently set to a Kronecker delta
distribution concentrating at $\ell(v_i)$; i.e., $(P_0)_{i} =
\delta_{\ell(v_i)}$.  Thus, on graphs with $V_U = \emptyset$ the
simplest \textsc{rw}-based information propagation is the \emph{label
  diffusion process} or simply \emph{diffusion process}
\begin{align}
\label{equ:diff}
  P_{t+1} \leftarrow T P_t,
\end{align}
where $(P_t)_i$ gives the distribution over $\ell(X_t^{(i)})$ at
iteration $t$.

Let $S \subseteq V$ be a set of nodes in $G$.  Given $T$ and $S$, we
define an \emph{absorbing random walk} to have the modified transition
probabilities $\hat{T}$, defined by:
\begin{align}
\label{equ:trans}
 \hat T_{ij} = \left\{
  \begin{array}{l l}
    0 &  \text{if $i \in S$ and $i\neq j$};\\
    1 &  \text{if $i \in S$ and $i = j$}; \\
    T_{ij} & \text{otherwise.}
  \end{array} \right.
\end{align}
Nodes in $S$ are ``absorbing'' in that a walk never leaves a node in
$S$ after it is encountered.  The $i$th row of $P_0$ now gives the
probability distribution for the first label encountered,
$\ell(X_0^{(i)})$, for an absorbing \textsc{rw} starting at $v_i$.  It
is easy to see by induction that by iterating the map
\begin{align}
\label{equ:lp}
  P_{t+1} \leftarrow \hat T P_t,
\end{align}
$(P_t)_i$ similarly gives the distribution over $\ell(X_t^{(i)})$.

In the case of partially labeled graphs we can now initialize the
label distributions for the unlabeled nodes $V_U$ with some prior, for
example a uniform distribution.\footnote{This prior could also be the
  output of an external classifier built on available node
  attributes.}  If we define the absorbing states to be the labeled
nodes, $S = V_L$, then the \emph{label propagation} algorithm
introduced in~\citep{ZhuGL03} can be cast in terms of simulating
label-absorbing \textsc{rw}s with transition probabilities given in
Eq.\ \eqref{equ:trans} until convergence, then assigning the most
probable absorbing label to the nodes in $V_U$.

The schemes discussed so far are two extreme cases of absorbing
\textsc{rw}s: one with no absorbing states, the diffusion process, and
one which absorbs at all labeled nodes, label propagation.  One useful
extension of absorbing \textsc{rw}s is to soften the definition of
absorbing states.  This can be naturally achieved by employing
\emph{partially absorbing random walks}~\citep{WU12}.  As the
propagation kernel framework does not require a specific propagation
scheme, we are free to choose any \textsc{rw}-based information
propagation scheme suitable for the given graph types. Based on the
basic techniques introduced in this section, we will suggest specific
propagation schemes for (un-)labeled, partially labeled, directed, and
attributed graphs as well as for graphs with regular structure in
Section~\ref{sec:Propagation_Schemes}.

\subsection{Steady-state Distributions vs. Early Stopping}
Assuming non-bipartite graphs, all \textsc{rw}s, absorbing or not,
converge to a steady-state distribution $P_\infty$
\citep{Lovasz96,WU12}.  Most existing \textsc{rw}-based approaches only
analyze the walks' steady-state distributions to make
predictions~\citep{KondorL02,ZhuGL03,WU12}.  However, \textsc{rw}s
without absorbing states converge to a constant steady-state
distribution, which is clearly uninteresting.  To address this, the
idea of \emph{early stopping} was successfully introduced into
power-iteration methods for clustering \citep{LinC10_PIC}, node-label
prediction~\citep{SzummerJ01}, as well as for the construction of a
kernel for node-label prediction \citep{NeumannGK13}.  The insight here
is that the intermediate distributions obtained by the \textsc{rw}s
during the convergence process provide useful insights about their
structure. In this paper, we adopt this idea for the construction of
graph kernels. That is, we use the entire evolution of distributions
encountered during \textsc{rw}s up to a given length to represent
graph structure.  This is accomplished by summing contributions
computed from the intermediate distributions of each iteration, rather
then only using the limiting distribution.  In the next section, we
define the family of propagation kernels.

\section{Propagation Kernel Framework}
\label{sec:PKs}
In this section, we introduce the general family of \emph{propagation
  kernels} (\textsc{pk}s).

\subsection{General Definition}
Here we will define a kernel $K\colon \cm{X} \times \cm{X} \to \bb{R}$
among graph instances $G^{(i)} \in \cm{X} $. The input space $\cm{X}$
comprises graphs $G^{(i)} = (V^{(i)},E^{(i)},\ell)$, where $V^{(i)}$
is the set of nodes and $E^{(i)}$ is the set of edges in graph
$G^{(i)}$. Edge weights are represented by weighted adjacency matrices
$A^{(i)} \in \bb{R}^{n_i \times n_i}$ and the label function~$\ell$
endows nodes with label and attribute information\footnote{Note that
  not both label and attribute information have to present and both
  could also be partially observed.}  as defined in the previous
section.

A simple way to compare two graphs $G^{(i)}$ and $G^{(j)}$ is to compare all pairs of nodes in
the two graphs:
\begin{align}
 K(G^{(i)}, G^{(j)}) =  \sum_{v \in G^{(i)}} \sum_{u \in G^{(j)}} k(u,v),  \notag
\end{align}
where $k(u,v)$ is an arbitrary node kernel determined by node labels
and, if present, node attributes.  This simple graph kernel, however,
does not account for graph structure given by the arrangement of node
labels and attributes in the graphs. Hence, we consider a
\emph{sequence} of graphs $G^{(i)}_t$ with evolving node information
based on information propagation, as introduced for node labels in the
previous section. We define the kernel contribution of iteration $t$
by
\begin{align}
 K(G^{(i)}_t, G^{(j)}_t) =  \sum_{v \in G^{(i)}_t} \sum_{u \in G^{(j)}_t} k(u,v).
 \label{equ:kernel_contib}
\end{align}
An important feature of propagation kernels is that the node kernel
$k(u, v)$ is defined in terms of the nodes' corresponding probability
distributions $p_{t,u}$ and $p_{t,v}$, which we update and maintain
throughout the process of information propagation.  For propagation
kernels between labeled and attributed graphs we define
\begin{align}
 k(u,v) = k_l(u,v) \cdot k_a(u,v),
 \label{equ:node_kernel}
\end{align}
where $k_l(u,v)$ is a kernel corresponding to label information and
$k_a(u,v)$ is a kernel corresponding to attribute information.  If no
attributes are present, then $k(u,v) = k_l(u,v)$.  The
$t_{\textsc{max}}$-iteration propagation kernel is now given by
\begin{align}
 K_{t_{\textsc{max}}}(G^{(i)}, G^{(j)}) = \sum_{t = 1}^{t_{\textsc{max}}} K(G^{(i)}_t, G^{(j)}_t).
 \label{equ:pk_kernel}
\end{align}

\begin{lem}
  \begin{center}
    Given that $k_l(u,v)$ and $k_a(u,v)$ are positive semidefinite
    node kernels, the propagation kernel $K_{t_{\textsc{max}}}$ is a
    positive semidefinite kernel.
  \end{center}
  \label{lem:psd}
\end{lem}
\textbf{Proof:} As $k_l(u,v)$ and $k_a(u,v)$ are assumed to be valid
node kernels, $k(u,v)$ is a valid node kernel as the product of
positive semidefinite kernels is again positive semidefinite. As for a
given graph $G^{(i)}$ the number of nodes is finite, $K(G^{(i)}_t,
G^{(j)}_t)$ is a convolution kernel \citep{Haussler99}.  As sums of
positive semidefinite matrices are again positive semidefinite, the
propagation kernel as defined in Eq.~\eqref{equ:pk_kernel} is positive
semidefinite.\\
\vspace{-0.8cm}\begin{flushright} $\Box$ \end{flushright}

\begin{algorithm}[t]
  \caption{The general propagation kernel computation.}
  \begin{algorithmic}
    \State \textbf{given:} graph database $\{G^{(i)}\}_i$, $\#$ iterations $t_{\textsc{max}}$, propagation scheme(s), base kernel $\langle \cdot, \cdot \rangle$
    \State $K \gets 0$, $initialize\;\, distributions\;\, P_0^{(i)}$
    \For{$t \gets 0\dotsc t_{\textsc{max}}$}
    \ForAll{graphs $G^{(i)}$}
    \ForAll{nodes $u \in G^{(i)}$}
    \State $quantize\;\;p_{t,u}$, where $p_{t,u} \text{ is } u\text{-th row in }P^{(i)}_t$ 			\Comment{bin node information}
    \EndFor
    \State $compute\;\;  \Phi_{i \cdot} = \phi(G^{(i)}_t)$  \Comment{count bin strengths}
    \EndFor
    \State $K \gets K + \langle \Phi, \Phi\rangle $ 	\Comment{compute and add kernel contribution}
    \ForAll{graphs $G^{(i)}$}
    \State $P^{(i)}_{t+1} \gets P^{(i)}_{t}$ 				\Comment{propagate node information}
    \EndFor
    \EndFor
  \end{algorithmic}
  \label{algo:propKernel}
\end{algorithm}

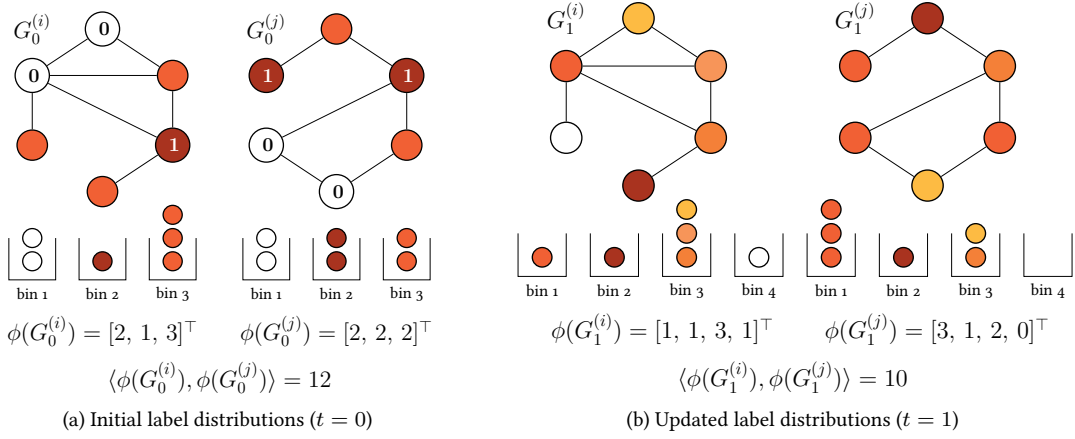
\begin{figure}[!t]
  \begin{center}
    \subfloat[Initial label distributions ($t=0$)]{\resizebox{0.41\textwidth}{!}{
        \begin{tikzpicture}
          [
            node_circle/.style={circle,thick,draw,text width=0.3cm,minimum size=6.5mm},
            bin_circle/.style={circle,thick,draw,text width=0.1cm},
            edge from parent/.style={very thick,draw=black!70}
          ]
          \begin{scope}
            \node (a1) [node_circle,fill=white] at (2,19.5) {\large$\bm{\,0}$};
            \node (a2) [node_circle,fill=white] at (0.5,18.5) {\large$\bm{\,0}$} edge [-] (a1);
            \node (a3) [node_circle,fill=RedOrange] at (0.5,17) {} edge [-] (a2);
            \node (a4) [node_circle,fill=Mahogany] at (3.5,17) {\large\color{white}$\bm{\,1}$} edge [-] (a2);
            \node (a5) [node_circle,fill=RedOrange] at (3.5,18.5) {} edge [-] (a1) edge [-] (a2) edge [-] (a4);
            \node (a6) [node_circle,fill=RedOrange] at (2,16) {} edge [-] (a4);
            \node (a) [bin_circle,fill=white] at (0.5,14.5) {};
            \node (a) [bin_circle,fill=white] at (0.5,15) {};
            \draw[color=black](0,15)--(0,14.1)--(1,14.1)--(1,15);
            \node at (0.5,13.8) {bin 1};
            \node (a) [bin_circle,fill=Mahogany] at (2,14.5) {};
            \draw[color=black](1.5,15)--(1.5,14.1)--(2.5,14.1)--(2.5,15);
            \node at (2,13.8) {bin 2};
            \node (a) [bin_circle,fill=RedOrange] at (3.5,14.5) {};
            \node (a) [bin_circle,fill=RedOrange] at (3.5,15) {};
            \node (a) [bin_circle,fill=RedOrange] at (3.5,15.5) {};
            \draw[color=black](3,15)--(3,14.1)--(4,14.1)--(4,15);
            \node at (3.5,13.8) {bin 3};
            \Large
            \node at (0.5,19.5) {$G^{(i)}_0$};
          \end{scope}
          \begin{scope}
            \node (b1) [node_circle,fill=RedOrange] at (7,19.5) {};
            \node (b2) [node_circle,fill=Mahogany] at (5.5,18.5) {\large\color{white}$\bm{\,1}$} edge [-] (b1);
            \node (b3) [node_circle,fill=white] at (5.5,17) {\large$\bm{\,0}$};
            \node (b4) [node_circle,fill=RedOrange] at (8.5,17) {};
            \node (b5) [node_circle,fill=Mahogany] at (8.5,18.5) {\large\color{white}$\bm{\,1}$} edge [-] (b1) edge [-] (b3) edge [-] (b4);
            \node (b6) [node_circle,fill=white] at (7,16) {\large$\bm{\,0}$} edge [-] (b3) edge [-] (b4);
            \node (b) [bin_circle,fill=white] at (5.5,14.5) {};
            \node (b) [bin_circle,fill=white] at (5.5,15) {};
            \draw[color=black](5,15)--(5,14.1)--(6,14.1)--(6,15);
            \node at (5.5,13.8) {bin 1};
            \node (b) [bin_circle,fill=Mahogany] at (7,14.5) {};
            \node (b) [bin_circle,fill=Mahogany] at (7,15) {};
            \draw[color=black](6.5,15)--(6.5,14.1)--(7.5,14.1)--(7.5,15);
            \node at (7,13.8) {bin 2};
            \node (b) [bin_circle,fill=RedOrange] at (8.5,14.5) {};
            \node (b) [bin_circle,fill=RedOrange] at (8.5,15) {};
            \draw[color=black](8,15)--(8,14.1)--(9,14.1)--(9,15);
            \node at (8.5,13.8) {bin 3};
            \Large
            \node at (5.5,19.5) {$G^{(j)}_0$};
          \end{scope}
          \node (b) [font=\Large] at (2,13) { $ \phi(G^{(i)}_0)=[2,\,1,\,3]\trans$ };
          \node (b) [font=\Large] at (7,13) { $ \phi(G^{(j)}_0)=[2,\,2,\,2]\trans$ };
          \node (b) [font=\Large] at (4.5,12) { $ \langle \phi(G^{(i)}_0),\phi(G^{(j)}_0)\rangle = 12$ };
      \end{tikzpicture}}}
      \hfill
      \subfloat[Updated label distributions ($t=1$)]{\resizebox{0.52\textwidth}{!}{
          \begin{tikzpicture}
            [
              node_circle/.style={circle,thick,draw,text width=0.3cm,minimum size=6.5mm},
              bin_circle/.style={circle,thick,draw,text width=0.1cm},
              edge from parent/.style={very thick,draw=black!70}
            ]

            \begin{scope}
              \node (a1) [node_circle,fill=Dandelion] at (2,9.5) {};
              \node (a2) [node_circle,fill=RedOrange] at (0.5,8.5) {} edge [-] (a1);
              \node (a3) [node_circle,fill=white] at (0.5,7) {} edge [-] (a2);
              \node (a4) [node_circle,fill=Orange] at (3.5,7) {} edge [-] (a2);
              \node (a5) [node_circle,fill=Peach] at (3.5,8.5) {} edge [-] (a1) edge [-] (a2) edge [-] (a4);
              \node (a6) [node_circle,fill=Mahogany] at (2,6) {} edge [-] (a4);
              \node (a) [bin_circle,fill=RedOrange] at (0,4.5) {};
              \draw[color=black](-0.5,5)--(-0.5,4.1)--(0.5,4.1)--(0.5,5);
              \node at (0,3.8) {bin 1};
              \node (a) [bin_circle,fill=Mahogany] at (1.5,4.5) {};
              \draw[color=black](1,5)--(1,4.1)--(2,4.1)--(2,5);
              \node at (1.5,3.8) {bin 2};
              \node (a) [bin_circle,fill=Orange] at (3,4.5) {};
              \node (a) [bin_circle,fill=Peach] at (3,5) {};
              \node (a) [bin_circle,fill=Dandelion] at (3,5.5) {};
              \draw[color=black](2.5,5)--(2.5,4.1)--(3.5,4.1)--(3.5,5);
              \node at (3,3.8) {bin 3};
              \node (a) [bin_circle,fill=white] at (4.5,4.5) {};
              \draw[color=black](4,5)--(4,4.1)--(5,4.1)--(5,5);
              \node at (4.5,3.8) {bin 4};
              \Large
              \node at (0.5,9.5) {$G^{(i)}_1$};
            \end{scope}

            \begin{scope}
              \node (b1) [node_circle,fill=Mahogany] at (8,9.5) {};
              \node (b2) [node_circle,fill=RedOrange] at (6.5,8.5) {} edge [-] (b1);
              \node (b3) [node_circle,fill=RedOrange] at (6.5,7) {};
              \node (b4) [node_circle,fill=RedOrange] at (9.5,7) {};
              \node (b5) [node_circle,fill=Orange] at (9.5,8.5) {} edge [-] (b1) edge [-] (b3) edge [-] (b4);
              \node (b6) [node_circle,fill=Dandelion] at (8,6) {} edge [-] (b3) edge [-] (b4);
              \node (b) [bin_circle,fill=RedOrange] at (6,4.5) {};
              \node (b) [bin_circle,fill=RedOrange] at (6,5) {};
              \node (b) [bin_circle,fill=RedOrange] at (6,5.5) {};
              \draw[color=black](5.5,5)--(5.5,4.1)--(6.5,4.1)--(6.5,5);
              \node at (6,3.8) {bin 1};
              \node (b) [bin_circle,fill=Mahogany] at (7.5,4.5) {};
              \draw[color=black](7,5)--(7,4.1)--(8,4.1)--(8,5);
              \node at (7.5,3.8) {bin 2};
              \node (b) [bin_circle,fill=Orange] at (9,4.5) {};
              \node (b) [bin_circle,fill=Dandelion] at (9,5) {};
              \draw[color=black](8.5,5)--(8.5,4.1)--(9.5,4.1)--(9.5,5);
              \node at (9,3.8) {bin 3};
              \draw[color=black](10,5)--(10,4.1)--(11,4.1)--(11,5);
              \node at (10.5,3.8) {bin 4};
              \Large
              \node at (6.5,9.5) {$G^{(j)}_1$};
            \end{scope}
            \node (b) [font=\Large] at (2.5,3) { $ \phi(G^{(i)}_1)=[1,\,1,\,3,\,1]\trans$ };
            \node (b) [font=\Large] at (8.2,3) { $ \phi(G^{(j)}_1)=[3,\,1,\,2,\,0]\trans$ };
            \node (b) [font=\Large] at (5.2,2) { $ \langle\phi(G^{(i)}_1),\phi(G^{(j)}_1)\rangle =10$ };

        \end{tikzpicture}}}
        \caption[]{\textbf{Propagation Kernel Computation.} Distributions,
          bins, count features, and kernel contributions for two graphs
          $G^{(i)}$ and $G^{(j)}$ with binary node labels and one
          iteration of label propagation, cf.\ Eq.~\eqref{equ:lp}, as the
          propagation scheme. Node-label distributions are decoded by
          color: white means $p_{0,u} = [1,0]$, dark red stands for
          $p_{0,u} = [0,1]$, and the initial distributions for unlabeled
          nodes (light red) are $p_{0,u} =
          [\nicefrac{1}{2},\nicefrac{1}{2}]$.}
        \label{fig:prop_kernel}
  \end{center}
\end{figure}

Let $|V^{(i)}| = n_i$ and $|V^{(j)}| = n_j$. Assuming that all node
information is given, the complexity of computing each contribution
between two graphs, Eq.~\eqref{equ:kernel_contib}, is $\cm{O}(n_i \,
n_j)$.  Even for medium-sized graphs this can be prohibitively
expensive considering that the computation has to be performed for
\emph{every} pair of graphs in a possibly large graph
database. However, if we have a node kernel of the form
\begin{align}
  k(u,v) = \left\{
  \begin{array}{l l}
    1\;\; &  \text{if } condition\\
    0\;\; &  \text{otherwise},
  \end{array} \right.
  \label{equ:dirac_kernel}
\end{align}
where $condition$ is an equality condition on the information of nodes
$u$ and $v$, we can compute $K$ efficiently by \emph{binning} the node
information, \emph{counting} the respective bin strengths for all
graphs, and \emph{computing a base kernel} among these counts.  That
is, we compute count features $\phi(G^{(i)}_t)$ for each graph and
plug them into a base kernel: $\langle \cdot,\cdot \rangle$
\begin{align}
 K(G^{(i)}_t, G^{(j)}_t) = \langle\phi(G^{(i)}_t),\phi(G^{(j)}_t)\rangle.
 \label{equ:feature_kernel}
\end{align}
In the simple case of a linear base kernel, the last step is just an
outer product of count vectors $\Phi_t \Phi_t^\top$, where the $i$th
row of $\Phi_t$, $\left( \Phi_t \right)_{i\, \cdot} =
\phi(G^{(i)}_t)$.  Now, for two graphs, binning and counting can be
done in $\cm{O}(n_i+n_j)$ and the computation of the linear base
kernel value is $\cm{O}(|\text{bins}|)$.  This is a commonly exploited
insight for efficient graph-kernel computation and it has already been
exploited for labeled graphs in previous
work~\citep{ShervashidzeB11,NeumannPGK12}.

Figure~\ref{fig:prop_kernel} illustrates the propagation kernel
computation for $t=0$ and $t=1$ for two example graphs and
Algorithm~\ref{algo:propKernel} summarizes the kernel computation for
a graph database $\mathbf{G} = \{G^{(i)}\}_i = (V,E,\ell)$ with a
total number of $N$ nodes.  From this general algorithm and
Eqs.~\eqref{equ:kernel_contib} and \eqref{equ:pk_kernel}, we see that
the two main components to design a propagation kernel are
\begin{itemize}
 \item the \textbf{node kernel} $k(u,v)$ comparing propagated
   information, and
 \item the \textbf{propagation scheme} $P^{(i)}_{t+1} \gets
   P^{(i)}_{t}$ propagating the information within the graphs.
\end{itemize}
The propagation scheme depends on the input graphs and we will give
specific suggestions for different graph types in
Section~\ref{sec:Propagation_Schemes}. Before defining the node
kernels depending on the available node information in
Section~\ref{sec:node_kernel}, we briefly discuss the general runtime
complexity of propagation kernels.

\subsection{Complexity Analysis}
The total runtime complexity of propagation kernels for a set of $n$
graphs with a total number of $N$ nodes and $M$ edges is
$\cm{O}\bigl((t_{\textsc{max}}-1) M + t_{\textsc{max}}
\,n^2\,n^{\star}\bigr)$, where $n^{\star} := \max_i(n_i)$.  For a pair
of graphs the runtime complexity of computing the count features, that
is, binning the node information and counting the bin strengths is
$\cm{O}(n_i + n_j)$.  Computing and adding the kernel contribution is
$\cm{O}(|\text{bins}|)$, where $|\text{bins}|$ is bounded by $n_i +
n_j$.  So, one iteration of the kernel computation for all graphs is
$\cm{O}(n^2\,n^{\star})$.  Note that in practice $|\text{bins}| \ll 2
n^{\star}$ as we aim to bin together similar nodes to derive a
meaningful feature representation.

Feature computation basically depends on propagating node information
along the edges of all graphs.  This operation depends on the number
of edges and the information propagated, so it is $\cm{O}((k+D)M) =
\cm{O}(M)$, where $k$ is the number of node labels and $D$ is the
attribute dimensionality. This operation has to be performed
$t_{\textsc{max}}-1$ times. Note that the number of edges is usually
much lower than $N^2$.

\section{Propagation Kernel Component 1: Node Kernel}
\label{sec:node_kernel}
In this section, we define node kernels comparing label distributions
and attribute information appropriate for the use in propagation
kernels. Moreover, we introduce \emph{locality sensitive hashing,}
which is used to discretize the distributions arsing from
\textsc{rw}-based information propagation and the continuous attribute
vectors directly.

\subsection{Definitions}
Above, we saw that one way to allow for efficient computation of
propagation kernels is to restrict the range of the node kernels to
$\lbrace 0, 1 \rbrace$.  Let us now define the two components of the
node kernel (Eq.~\eqref{equ:node_kernel}) in this form.  The
\emph{label kernel} can be represented as
\begin{align}
  k_l(u,v) = \left\{
  \begin{array}{l l}
    1\;\; &  \text{if } h_l(p_{t,u}) = h_l(p_{t,v});\\
    0\;\; &  \text{otherwise,}
  \end{array} \right.
\label{equ:label_kernel}
\end{align}
where $p_{t,u}$ is the node-label distribution of node $u$ at
iteration $t$ and $h_l(\cdot)$ is a quantization
function~\citep{Gersho1991}, more precisely a locality sensitive hash
(\textsc{lsh}) function~\citep{pstable}, which will be introduced in
more detail in the next section.  Note that $p_{t,u}$ denotes a row in
the label distribution matrix $P^{(i)}_t$, namely the row
corresponding to node $u$ of graph $G^{(i)}$.

\begin{figure}[t]
\begin{center}
\subfloat[$k_a(u,v)$\hspace*{-0.35cm}]{\includegraphics[width=0.3\textwidth]{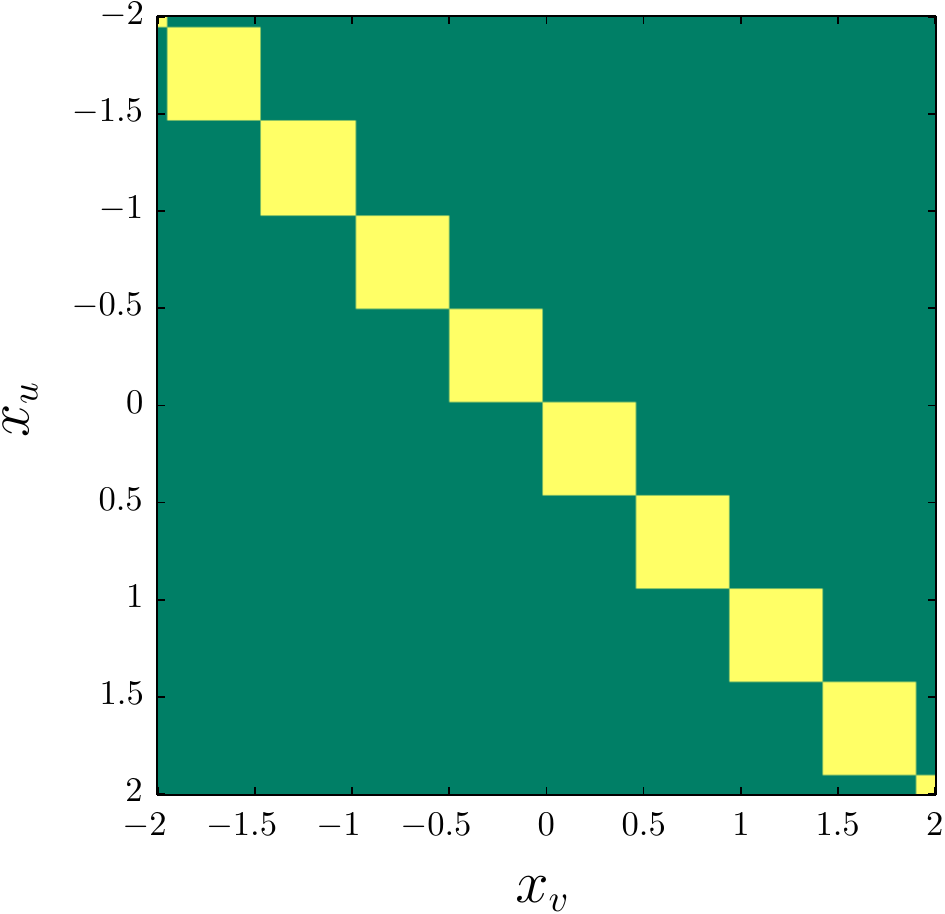}\label{fig:node_kernel_hash}}
\hfill
\subfloat[thresholded Gaussian\hspace*{-0.4cm}]{\includegraphics[width=0.3\textwidth]{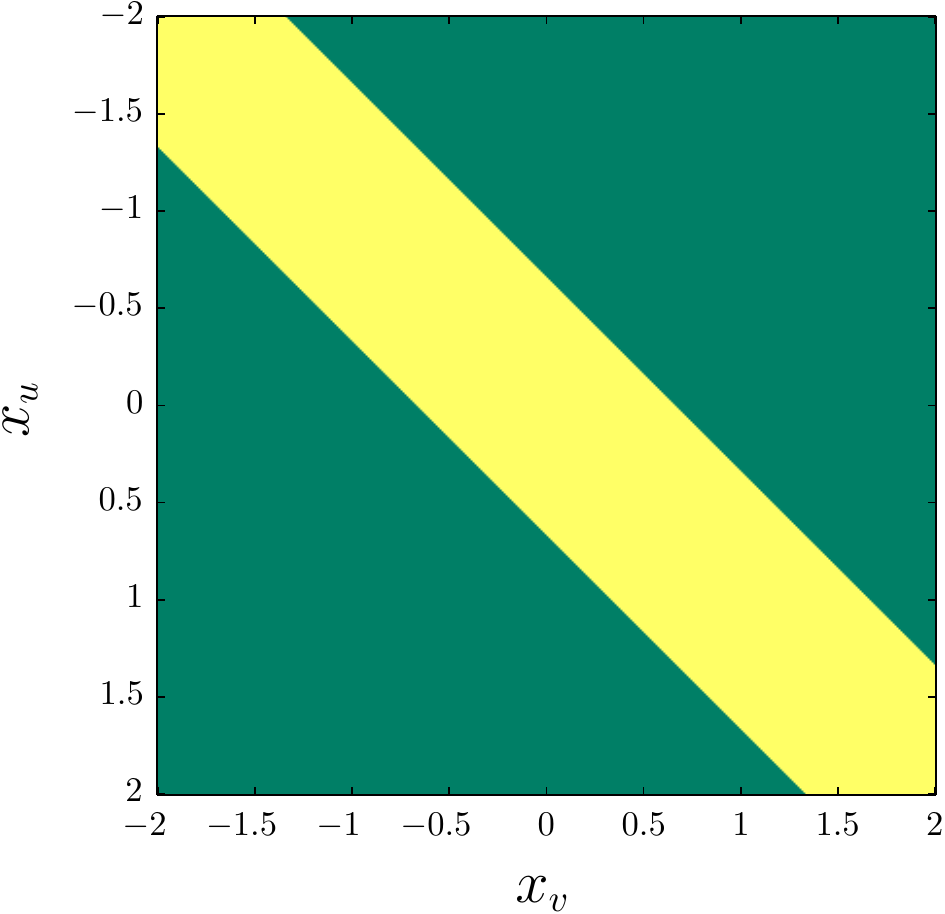}\label{fig:node_kernel_thres}}
\hfill
\subfloat[Gaussian kernel]{\includegraphics[width=0.352\textwidth]{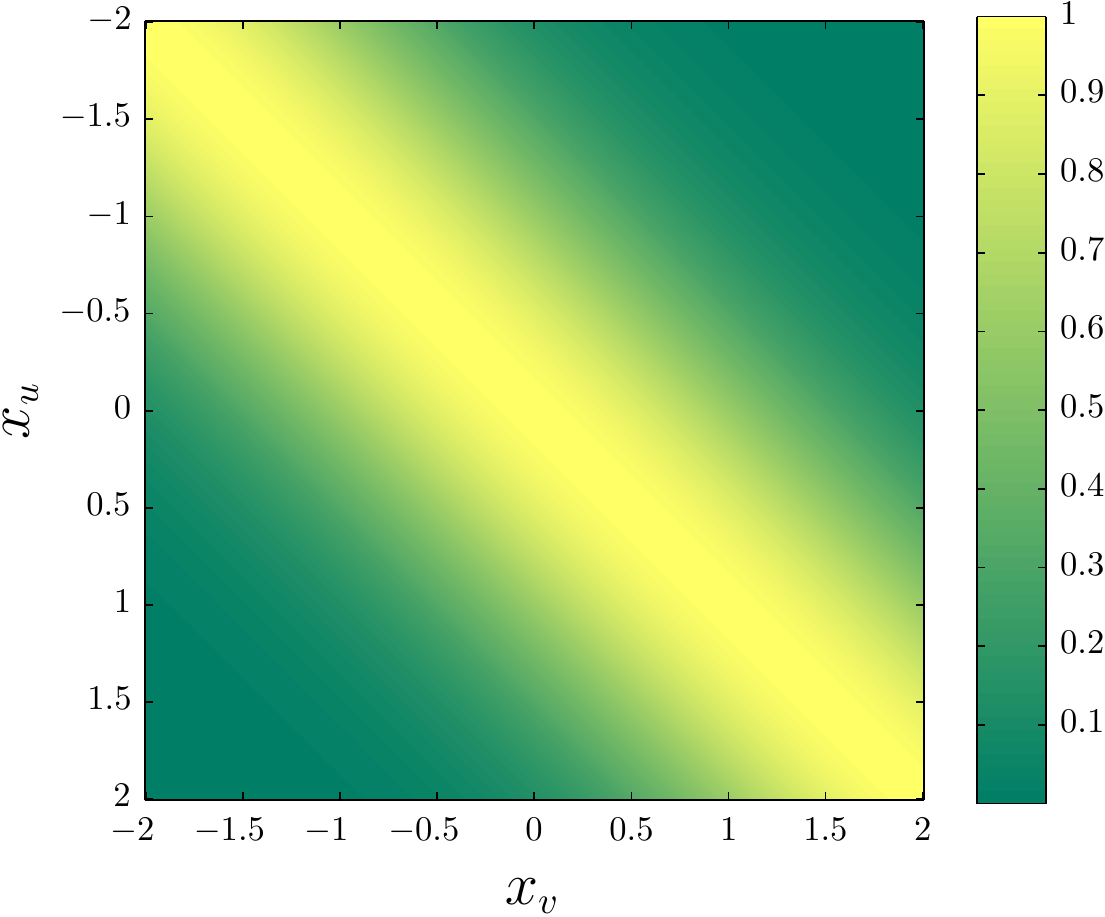}\label{fig:node_kernel_gauss}}
\caption[]{\textbf{Attribute Kernels.} Three different attribute
  functions among nodes with continuous attributes $x_u$ and $x_v \in
  \left[ -2,2 \right]$. Panel (a) illustrates an attribute kernel
  suitable for the efficient computation in propagation kernels, Panel
  (b) shows a thresholded Gaussian, and Panel (c) a Gaussian kernel. }
\label{fig:node_kernels}
\end{center}
\end{figure}

Propagation kernels can be computed for various kinds of attribute
kernels as long as they have the form of Eq.~\eqref{equ:dirac_kernel}.
The most rudimentary \emph{attribute kernel} is
\begin{align}
  k_{a}(u,v) = \left\{
  \begin{array}{l l}
    1\;\; &  \text{if } h_{a}(x_u) = h_{a}(x_v)\\
    0\;\; &  \text{otherwise},
  \end{array} \right.
  \label{equ:simple_attr_kernel}
\end{align}
where $x_u$ is the one-dimensional continuous attribute of node $u$
and $h_{a}(\cdot)$ is again an \textsc{lsh} function.
Figure~\ref{fig:node_kernels} contrasts this simple attribute kernel
for a one-dimensional attribute to a thresholded Gaussian function and
the Gaussian kernel commonly used to compare node attributes in graph
kernels.  To deal with higher-dimensional attributes, we can choose
the attribute kernel to be the product of kernels on each attribute
dimension:
\begin{align}
  k_a(u,v) = \prod_{d=1}^D k_{a_d}(u,v), \text{ where}  \notag \\
  k_{a_d}(u,v) = \left\{
  \begin{array}{l l}
    1\;\; &  \text{if } h_{a_d}(x^{(d)}_u) = h_{a_d}(x^{(d)}_v);\\
    0\;\; &  \text{otherwise},
  \end{array} \right.
\label{equ:attr_kernel}
\end{align}
where $x^{(d)}_u$ is the respective dimension of the attribute
$\mathbf{x}_u$ of node $u$.  Note that each dimension now has its own
\textsc{lsh} function $h_{a_d}(\cdot)$.  However, analogous to the
label kernel, we can also define an attribute kernel based on
propagated attribute distributions
\begin{align}
  k_{a}(u,v) = \left\{
  \begin{array}{l l}
    1\;\; &  \text{if } h_{a}(q_{t,u}) = h_{a}(q_{t,v})\\
    0\;\; &  \text{otherwise},
  \end{array} \right.
\label{equ:attr_distr_kernel}
\end{align}
where $q_{t,u}$ is the attribute distribution of node $u$ at iteration
$t$. Next we explain the locality sensitive hashing approach used to
discretize distributions and continuous attributes. In
Section~\ref{sec:attributes}, we will then derive an efficient way to
propagate and hash continuous attribute distributions.

\subsection{Locality Sensitive Hashing}
We now describe our quantization approach for implementing propagation
kernels for graphs with node-label distributions and continuous
attributes.  The idea is inspired by locality-sensitive
hashing~\citep{pstable} which seeks quantization functions on metric
spaces where points ``close enough'' to each other in that space are
``probably'' assigned to the same bin.  In the case of distributions,
we will consider each node-label vector as being an element of the
space of discrete probability distributions on $k$ items equipped with
an appropriate probability metric.  If we want to hash attributes
directly, we simply consider metrics for continuous values.

\begin{my_def}[Locality Sensitive Hash (\textsc{lsh})]
Let $\cm{X}$ be a metric space with metric $d\colon \cm{X} \times
\cm{X} \to \mathbb{R}$, and let $\cm{Y} = \lbrace 1, 2, \dotsc, k'
\rbrace$. Let $\theta > 0$ be a threshold, $c > 1$ be an approximation
factor, and $p_1, p_2 \in (0, 1)$ be the given success probabilities.
A set of functions $\cm{H}$ from $\cm{X}$ to $\cm{Y}$ is called a
$(\theta, c\theta, p_1, p_2)$-\emph{locality sensitive hash} if for
any function $h \in \cm{H}$ chosen uniformly at random, and for any
two points $x, x' \in \cm{X}$, it holds that\vspace{0.15cm} \\
\hspace*{2mm}$-$ if $d(x, x') < \phantom{c}\theta$, then $\Pr(h(x) = h(x')) > p_1$, and \\
\hspace*{2mm}$-$  if $d(x, x') > c\theta$, then $\Pr(h(x) = h(x')) < p_2$.
\label{def:lsh}
\end{my_def}
It is known that we can construct \textsc{lsh} families for $\ell^p$ spaces
with $p \in (0, 2]$~\citep{pstable}.  Let $V$ be a real-valued random
variable.  $V$ is called $p$\emph{-stable} if for any $\lbrace x_1,
x_2,$ $\dotsc, x_d \rbrace$, $x_i \in \mathbb{R}$ and independently
sampled $v_1, v_2, \dotsc, v_d$, we have $\sum x_i v_i \sim \lVert
\bm{x} \rVert_p V$.  Explicit $p$-stable distributions are known for
some $p$; for example, the standard Cauchy distribution is 1-stable,
and the standard normal distribution is 2-stable.  Given the ability
to sample from a $p$-stable distribution $V$, we may define an
\textsc{lsh} $\cm{H}$ on $\mathbb{R}^d$ with the $\ell^p$ metric
\citep{pstable}. An element $h$ of $\cm{H}$ is specified by three
parameters: a width $w \in \mathbb{R}^+$, a $d$-dimensional vector
$\bm{v}$ whose entries are independent samples of $V$, and $b$ drawn
from $\cm{U}[0, w]$. Given these, $h$ is then defined as
\begin{equation}\label{lsh_definition}
  h(\bm{x}; w, \bm{v}, b) =
  \left \lfloor
  \frac{\bm{v}\trans \bm{x} + b}{w}
  \right \rfloor.
\end{equation}
We may now consider $h(\cdot)$ to be a function mapping our
distributions or attribute values to integer-valued bins, where
similar distributions end up in the same bin.  Hence, we obtain node
kernels as defined in Eqs.~\eqref{equ:label_kernel} and
\eqref{equ:attr_distr_kernel} in the case of distributions, as well as
simple attribute kernels as defined in
Eqs.~\eqref{equ:simple_attr_kernel} and \eqref{equ:attr_kernel}.  To
decrease the probability of collision, it is common to choose more
than one random vector $\bm{v}$.  For propagation kernels, however, we
only use one hyperplane, as we effectively have $t_{\textsc{max}}$
hyperplanes for the whole kernel computation and the probability of a
hash conflict is reduced over the iterations.

\begin{algorithm}[t]
  \caption{\textsc{calculate-lsh}}
  \begin{algorithmic}
    \State \textbf{given:} matrix $X \in \bb{R}^{N \times D}$, bin width $w$, metric \textsc{m}
    \If {\textsc{m} = \textsc{h}}
      \State $X \gets \sqrt{X}$ 				\Comment{square root transformation}
    \EndIf
    \If {\textsc{m} = \textsc{h} or \textsc{m} = \textsc{l2}} \Comment{generate random projection vector}
	\State $v \gets \textsc{rand-norm}(D)$ 			\Comment{sample from $\cm{N}(0,1)$}
    \ElsIf { \textsc{m} = \textsc{tv} or \textsc{m} = \textsc{l1}}
      \State $v \gets \textsc{rand-norm}(D) / \textsc{rand-norm}(D)$ 		\Comment{sample from $\cm{C}auchy(0,1)$}
    \EndIf
    \State $b = w * \text{\textsc{rand-unif()}}$				\Comment{random offset $b \sim \cm{U}\left[0,w\right]$ }
    \State $h(X) =  \text{floor}((X*v + b) / w)$		\Comment{compute hashes}
  \end{algorithmic}
  \label{algo:LSH}
\end{algorithm}

The intuition behind the expression in Eq.~\eqref{lsh_definition} is
that $p$-stability implies that two vectors that are close under the
$\ell^p$ norm will be close after taking the dot product with
$\bm{v}$; specifically, $(\bm{v}\trans \bm{x} - \bm{v}\trans \bm{x'})$
is distributed as $\lVert \bm{x} - \bm{x}' \rVert_p V$.  So, in the
case where we want to construct a hashing for $D$-dimensional
continuous node attributes to preserve $\ell^1$ (\textsc{l1}) or
$\ell^2$ (\textsc{l2}) distance
\begin{align*}
  d_{\text{\textsc{l1}}}(x_u, x_v)
  =
  \sum^D_{d=1}\, \lvert x^{(d)}_u - x^{(d)}_v \rvert, \;\;\;
  d_{\text{\textsc{l2}}}(x_u, x_v)
  =
  \left (
  \sum^D_{d=1}\,
  \bigl(x^{(d)}_u - x^{(d)}_v\bigr)^2
  \right)^{\nicefrac{1}{2}}\!\!\!\!,
\end{align*}
we directly apply Eq.~\eqref{lsh_definition}.  In the case of
distributions, we are concerned with the space of discrete probability
distributions on $k$ elements, endowed with a probability metric
$d$. Here we specifically consider the \emph{total variation}
(\textsc{tv}) and \emph{Hellinger} (\textsc{h}) distances:
\begin{align*}
  d_{\text{\textsc{tv}}}(p_u, p_v)
  =
  \nicefrac{1}{2} \sum^k_{i=1}\, \lvert p^{(i)}_u - p^{(i)}_v \rvert, \;\;\;
  d_{\text{\textsc{h}}}(p_u, p_v)
  =
  \left (
  \nicefrac{1}{2}
  \sum^k_{i=1}\,
  \bigl(\sqrt{p^{(i)}_u} - \sqrt{p^{(i)}_v}\bigr)^2
  \right)^{\nicefrac{1}{2}}\!\!\!\!.
\end{align*}
The total variation distance is simply half the $\ell^1$ metric, and
the Hellinger distance is a scaled version of the $\ell^2$ metric
after applying the map $p \mapsto \sqrt{p}$.  We may therefore create
a locality-sensitive hash family for $d_{\text{\textsc{tv}}}$ by
direct application of Eq.~\eqref{lsh_definition} and create a
locality-sensitive hash family for $d_{\text{\textsc{h}}}$ by using
Eq.~\eqref{lsh_definition} after applying the square root map to our
label distributions. The \textsc{lsh} computation for a matrix $X \in
\bb{R}^{N \times D}$, where the $u$-th row of $X$ is $\bm{x}_u$, is
summarized in Algorithm~\ref{algo:LSH}.

\section{Propagation Kernel Component 2: Propagation Scheme}
\label{sec:Propagation_Schemes}
As pointed out in the introduction, the input graphs for graph kernels
may vary considerably. One key insight to the design of efficient and
powerful propagation kernels is to choose an appropriate propagation
scheme for the graph dataset at hand.  By utilizing random walks
(\textsc{rw}s) we are able to use efficient off-the-shelf algorithms,
such as label diffusion or label
propagation~\citep{SzummerJ01,ZhuGL03,WU12}, to implement information
propagation on the input graphs.  In this section, we explicitly
define propagation kernels appropriate for fully labeled, unlabeled,
partially labeled, directed, and attributed graphs as well as for
graphs with a regular grid structure.  In each particular algorithm,
the specific parts changing compared to the general propagation kernel
computation (Algorithm~\ref{algo:propKernel}) will be marked in
color.

\subsection{Labeled and Unlabeled Graphs}
\label{sec:PK_labeledGraphs}
For \emph{fully labeled graphs} we suggest the use of the label
diffusion process from Eq.~\eqref{equ:diff} as the propagation
scheme. Given a database of fully labeled graphs $\mathbf{G} =
\{G^{(i)}\}_{i=1,\dots,n}$ with a total number of $N=\sum_i n_i$
nodes, label diffusion on all graphs can be efficiently implemented by
multiplying a sparse block-diagonal transition matrix $T \in
\bb{R}^{N\times N}$, where the blocks are the transition matrices
$T^{(i)}$ of the respective graphs, with the label distribution matrix
$P_t = \left[P^{(1)}_t,\dots,P^{(n)}_t \right]^\top \in
\bb{R}^{N\times k}$. This can be done efficiently due to the sparsity
of $T$.  The propagation kernel computation for labeled graphs is
summarized in Algorithm~\ref{algo:propKernel_labeled}. The specific
parts compared to the general propagation kernel computation
(Algorithm~\ref{algo:propKernel}) for fully labeled graphs are marked
in green (input) and blue (computation).  For \emph{unlabeled graphs}
we suggest to set the label function to be the node degree $\ell(u) =
\text{degree}(u)$ and then apply the same \textsc{pk} computation as
for fully labeled graphs.

\begin{algorithm}[t]
  \caption{Propagation kernel for fully labeled graphs.}
  \begin{algorithmic}
    \State \textbf{given:} graph database $G = (V,E,\ell)$, $\#$ iterations $t_{\textsc{max}}$, \textcolor{ForestGreen}{transition matrix} $\color{ForestGreen} T$, bin width $w$, metric \textsc{m}, base kernel $\langle \cdot, \cdot \rangle$
    \State \textbf{initialization:} $K \gets 0$, $\color{blue} P_0 \gets \delta_{\ell(V)}$
    \For{$t \gets 0\dotsc t_{\textsc{max}}$}
    \State \textsc{calculate-lsh}($P_t$, $w$, \textsc{m}) 	\Comment{bin node information}
    \ForAll{graphs $G^{(i)}$}
    \State $compute\;\;  \Phi_{i \cdot} = \phi(G^{(i)}_t)$  \Comment{count bin strengths}
    \EndFor

    \State $\color{blue} P_{t+1} \gets T P_{t}$ 				\Comment{\textcolor{blue}{label diffusion}}
    \State $K \gets K + \langle \Phi, \Phi\rangle $ 	\Comment{compute and add kernel contribution}
    \EndFor

  \end{algorithmic}
  \label{algo:propKernel_labeled}
\end{algorithm}

\subsection{Partially Labeled and Directed Graphs}
For \emph{partially labeled graphs,} where some of the node labels are
unknown, we suggest \emph{label propagation} as an appropriate
propagation scheme.  Label propagation differs from label diffusion in
the fact that before each iteration of the information propagation,
the labels of the originally labeled nodes are pushed
back~\citep{ZhuGL03}. Let $P_0 = \left[ P_{0,[\text{labeled}]},
  P_{0,[\text{unlabeled}]}\right]\trans$ represent the prior label
distributions for the nodes of all graphs in a graph database
$\mathbf{G} = (V,E,\ell)$, where the distributions in
$P_{0,[\text{labeled}]}$ represent observed labels and
$P_{0,[\text{unlabeled}]}$ are initialized uniformly. Then label
propagation is defined by
\begin{align}
  P_{t,[labeled]} &\leftarrow P_{0,[\text{labeled}]}; \notag \\
  P_{t+1}  &\leftarrow T \, P_t.
  \label{equ:LP}
\end{align}
Note that this propagation scheme is equivalent to the one defined in
Eq.~\eqref{equ:lp} using partially absorbing \textsc{rw}s.  Other
similar update schemes, such as ``label spreading''~\citep{ZhouBLWS03},
could be used in a propagation kernel as well. Thus, the propagation
kernel computation for partially labeled graphs is essentially the
same as Algorithm~\ref{algo:propKernel_labeled}, where the
initialization for the unlabeled nodes has to be adapted, and the
(partial) label push back has to be added before the node information
is propagated. The relevant parts are the ones marked in blue.  Note
that for graphs with large fractions of labeled nodes it might be
preferable to use label diffusion even though they are partially
labeled.

To implement propagation kernels between \emph{directed graphs}, we
can proceed as above after simply deriving transition matrices
computed from the potentially non-symmetric adjacency matrices. That
is, for the propagation kernel computation only the input changes
(marked in green in Algorithm~\ref{algo:propKernel_labeled}).  The
same idea allows weighted edges to be accommodated; again, only the
transition matrix has to be adapted.  Obviously, we can also combine
partially labeled graphs with directed or weighted edges by changing
both the blue and green marked parts accordingly.

\subsection{Graphs with Continuous Node Attributes}
\label{sec:attributes}
Nowadays, learning tasks often involve graphs whose nodes are
attributed with continuous information.  Chemical compounds can be
annotated with the length of the secondary structure elements (the
nodes) or measurements for various properties, such as hydrophobicity
or polarity. 3\textsc{d} point clouds can be enriched with curvature
information, and images are inherently composed of 3-channel color
information. All this information can be modeled by continuous node
attributes. In Eq.~\eqref{equ:simple_attr_kernel} we introduced a
simple way to deal with attributes. The resulting propagation kernel
essentially counts similar label arrangements only if the
corresponding node attributes are similar as well.  Note that for
higher-dimensional attributes it can be advantageous to compute
separate \textsc{lsh}s per dimension, leading to the node kernel
introduced in Eq.~\eqref{equ:attr_kernel}. This has the advantage that
if we standardize the attributes, we can use the same bin-width
parameter $w_a$ for all dimensions.  In all our experiments we
normalize each attribute to have unit standard deviation and will set
$w_a = 1$.  The disadvantage of this method, however, is that the
arrangement of attributes in the graphs is ignored.

\begin{algorithm}[t]
  \caption{Propagation kernel (\textsc{p2k}) for attributed graphs.}
  \begin{algorithmic}
    \State \textbf{given:}  graph database $G = (V,E,\ell)$, $\#$ iterations $t_{\textsc{max}}$, \textcolor{ForestGreen}{transition matrix} $\color{ForestGreen} T$,
			    bin widths $w_l$, $w_a$, metrics \textsc{m}$_l$, \textsc{m}$_a$, base kernel $\langle \cdot, \cdot \rangle$
    \State \textbf{initialization:} $K \gets 0$, $P_0 \gets \delta_{\ell(V)}$
    \State $\color{blue}\boldsymbol{\mu} = X$, $\color{blue}\Sigma = \cov(X)$,  $\color{blue}W_{0}=I$	\Comment{\textcolor{blue}{\textsc{gm} initialization}}
    \State $\color{blue}y \gets \text{\textsc{rand}}(\text{num-samples})$				\Comment{\textcolor{blue}{sample points for \textsc{gm} evaluations}}
    \For{$t \gets 0\dotsc t_{\textsc{max}}$}
      \State $h_l \gets$ \textsc{calculate-lsh}($P_t$, $w_l$, \textsc{m}$_l$) 		\Comment{bin label distributions}
      \State $\color{blue} Q_t \gets$ \textcolor{blue}{\textsc{evaluate-pdfs}($\boldsymbol{\mu}$, $\Sigma$, $W_{t}$, $y$)} 	\Comment{\textcolor{blue}{evaluate \textsc{gm}s at $y$}}
      \State $\color{blue} h_a \gets$ \textcolor{blue}{\textsc{calculate-lsh}($Q_t$, $w_a$, \textsc{m}$_a$)} 		\Comment{\textcolor{blue}{bin attribute distributions}}
      \State $\color{blue} h \gets h_l \wedge h_a$							\Comment{\textcolor{blue}{combine label and attribute bins}}
      \ForAll{graphs $G^{(i)}$}
	  \State $compute\;\;  \Phi_{i \cdot} = \phi(G^{(i)}_t)$  \Comment{count bin strengths}
      \EndFor
      \State $P_{t+1} \gets T P_{t}$ 				\Comment{propagate label information}
      \State $\color{blue} W_{t+1} \gets T W_{t}$ 				\Comment{\textcolor{blue}{propagate attribute information}}
      \State $K \gets K + \langle \Phi, \Phi\rangle $ 	\Comment{compute and add kernel contribution}
    \EndFor
  \end{algorithmic}
  \label{algo:p2k}
\end{algorithm}

In the following, we derive \textsc{p2k,} a variant of propagation
kernels for \emph{attributed graphs} based on the idea of propagating
both attributes and labels. That is, we model graph similarity by
comparing the arrangement of labels \emph{and} the arrangement of
attributes in the graph. The attribute kernel for \textsc{p2k} is
defined as in Eq.~\eqref{equ:attr_distr_kernel}; now the question is
how to efficiently propagate the continuous attributes and how to
efficiently model and hash the distributions of (multivariate)
continuous variables. Let $X \in \bb{R}^{N \times D}$ be the design
matrix, where the $u$th row in $X$ is the attribute vector of node
$u$, $\mathbf{x}_u$.  We will associate with each node of each graph a
probability distribution defined on the attribute space, $q_u$, and
will update these as attribute information is propagated across graph
edges as before.  One challenge in doing so is ensuring that these
distributions can be represented with a finite description.  The
discrete label distributions from before were naturally finite
dimensional and could be compactly represented and updated via the
$P_t$ matrices.  We seek a similar representation for attributes.  Our
proposal is to define the node-attribute distributions to be mixtures
of $D$-dimensional multivariate Gaussians, one centered on each
attribute vector in $X$:
\begin{equation*}
  q_u = \sum_{v} W_{uv}\, \cm{N}(\mathbf{x}_v, \Sigma),
\end{equation*}
where the sum ranges over all nodes $v$, $W_{u, \cdot}$ is a vector of
mixture weights, and $\Sigma$ is a shared $D \times D$ covariance
matrix for each component of the mixture.  In particular, here we set
$\Sigma$ to be the sample covariance matrix calculated from the $N$
vectors in $X$.  Now the $N \times N$ row-normalized $W$ matrix can be
used to compactly represent the entire set of attribute distributions.
As before, we will use the graph structure to iteratively spread
attribute information, updating these $W$ matrices, deriving a
sequence of attribute distributions for each node to use as inputs to
node attribute kernels in a propagation kernel scheme.

We begin by defining the initial weight matrix $W_0$ to be the
identity matrix; this is equivalent to beginning with each node
attribute distribution being a single Gaussian centered on the
corresponding attribute vector:
\begin{align*}
  W_0 &= I; \notag \\
  q_{0,u} &= \cm{N}(\mathbf{x}_u ,\Sigma). \notag
\end{align*}
Now, in each propagation step the attribute distributions are updated
by the distribution of their neighboring nodes $Q_{t+1} \gets Q_t$.
We accomplish this by propagating the mixture weights $W$ across the
edges of the graph according to a row-normalized transition matrix
$T$, derived as before:
\begin{align}
  W_{t+1} &\gets T W_{t} = T^t; \notag \\
  q_{t+1,u} &= \sum_v \left( W_{t}\right) _{uv}\, \cm{N}(\mathbf{x}_v ,\Sigma).
\label{equ:mixture_approx}
\end{align}

We have described how we associate attribute distributions with each
node and how we update them via propagating their weights across the
edges of the graph.  However, the weight vectors contained in $W$ are
not themselves directly suitable for comparing in an attribute kernel
$k_a$, because any information about the similarity of the mean
vectors is ignored.  For example, imagine that two nodes $u$ and $v$
had exactly the same attribute vector, $\mathbf{x}_u = \mathbf{x}_v$.
Then mass on the $u$ component of the Gaussian mixture is exchangeable
with mass on the $v$ component, but typical vectorial kernels cannot
capture this.  For this reason, we use our Gaussian mixtures to
associate a vector more appropriate for kernel comparison with each
node.  Namely, we select a fixed set of sample points in attribute
space (in our case, chosen uniformly from the node attribute vectors
in $X$), evaluate the \textsc{pdf}s associated with each node at these
points, and use this vector to summarize the nodes.  This handles the
exchangeability issue from above and also allows a more compact
representation for hash inputs; in our experiments, we used 100 sample
points and achieved good performance.  As before, these vectors can
then be hashed jointly or individually for each sample point. Note
that the bin width $w_a$ has to be adapted accordingly. In our
experiments, we will use the latter option and set $w_a = 1$ for all
datasets.  The computational details of \textsc{p2k} are given in
Algorithm~\ref{algo:p2k}, where the additional parts compared to
Algorithm~\ref{algo:propKernel} are marked in blue (computation) and
green (input).  An extension to Algorithm~\ref{algo:p2k} would be to
refit the \textsc{gm}s after a couple of propagation iterations.  We
did not consider refitting in our experiments as the number of kernel
iterations $t_{\textsc{max}}$ was set to $10$ or $15$ for all datasets
-- following the descriptions in existing work on iterative graph
kernels \citep{ShervashidzeB11,NeumannPGK12}.

\subsection{Grid Graphs}
\label{sec:PK_gridGraphs}
One of our goals in this paper is to compute propagation kernels for
pixel grid graphs.  A graph kernel among grid graphs can be defined
such that two grids should have a high kernel value if they have
similarly arranged node information. This can be naturally captured by
propagation kernels as they monitor information spread on the grids.
Na\"{i}vely, one could think that we can simply apply
Algorithm~\ref{algo:propKernel_labeled} to achieve this goal.
However, given that the space complexity of this algorithm scales with
the number of edges and even medium sized images such as texture
patches will easily contain thousands of nodes, this is not feasible.
For example considering $100 \times 100$-pixel image patches with an
8-neighborhood graph structure, the space complexity required would be
$2.4$ million units\footnote{Using a coordinate list sparse
  representation, the memory usage per pixel grid graph for
  Algorithm~\ref{algo:propKernel_labeled} is $\cm O(3m_1m_2p)$, where
  $m_1 \times m_2$ are the grid dimensions and p is the size of the
  pixel neighborhood.} (floating point numbers) \emph{per graph}.
Fortunately, we can exploit the flexibility of propagation kernels by
exchanging the propagation scheme.  Rather than label diffusion as
used earlier, we employ discrete convolution; this idea was introduced
for efficient clustering on discrete lattices in \citep{BauckhageK13}.
In fact, isotropic diffusion for denoising or sharpening is a highly
developed technique in image processing \citep{Jahne05}. In each
iteration, the diffused image is derived as the convolution of the
previous image and an isotropic (linear and space-invariant) filter.
In the following, we derive a space- and time-efficient way of
computing propagation kernels for grid graphs by means of
convolutions.

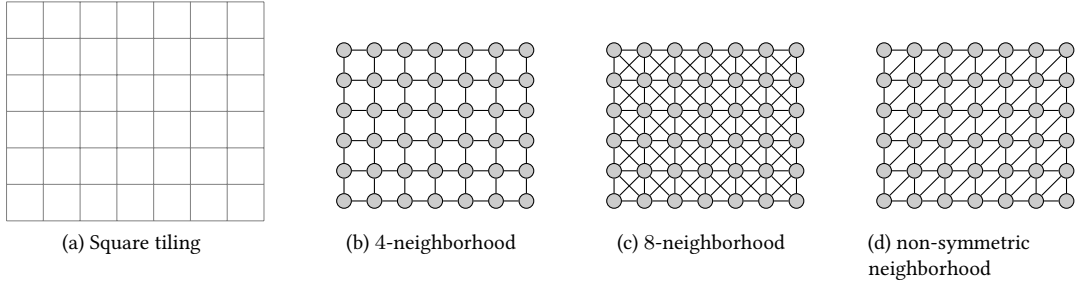
\begin{figure}[!t]
\subfloat[Square tiling]{
 \resizebox{0.24\textwidth}{!}{\begin{tikzpicture} \draw[step=0.5cm,gray,very thin] (0,0) grid (3.5,3);\end{tikzpicture}}}
\hfill
\subfloat[$4$-neighborhood]{
  \begin{tikzpicture}[darkstyle/.style={circle,draw,fill=gray!40,inner sep=0pt,minimum size=5.5}]
    \foreach \x in {0,...,6}
      \foreach \y in {0,...,5}{
	\node [darkstyle]  (\x\y) at (0.4*\x,0.4*\y) {};}
    \foreach \x in {0,...,6}
      \foreach \y [count=\yi] in {0,...,4}
	\draw (\x\y)--(\x\yi);
    \foreach \x [count=\xi] in {0,...,5}
      \foreach \y in {0,...,5}
	\draw (\x\y)--(\xi\y) ;
    \node (dummy) at (0,-0.15) {};
  \end{tikzpicture}}\hfill
\subfloat[$8$-neighborhood]{
  \begin{tikzpicture}[darkstyle/.style={circle,draw,fill=gray!40,inner sep=0pt,minimum size=5.5}]
    \foreach \x in {0,...,6}
      \foreach \y in {0,...,5}{
	\node [darkstyle]  (\x\y) at (0.4*\x,0.4*\y) {};}
    \foreach \x in {0,...,6}
      \foreach \y [count=\yi] in {0,...,4}
	\draw (\x\y)--(\x\yi);
    \foreach \x [count=\xi] in {0,...,5}
      \foreach \y in {0,...,5}
	\draw (\x\y)--(\xi\y) ;
    \foreach \x [count=\xi] in {0,...,5}
      \foreach \y [count=\yi] in {0,...,4}
	\draw (\x\y)--(\xi\yi);
    \foreach \x [count=\xi] in {0,...,5}
      \foreach \y [count=\yi] in {0,...,4}
	\draw (\xi\y)--(\x\yi);
    \node (dummy) at (0,-0.15) {};
  \end{tikzpicture}} \hfill
\subfloat[non-symmetric, neighborhood][non-symmetric\\ neighborhood]{
  \begin{tikzpicture}[darkstyle/.style={circle,draw,fill=gray!40,inner sep=0pt,minimum size=5.5}]
    \foreach \x in {0,...,6}
      \foreach \y in {0,...,5}{
	\node [darkstyle]  (\x\y) at (0.4*\x,0.4*\y) {};}
    \foreach \x in {0,...,6}
      \foreach \y [count=\yi] in {0,...,4}
	\draw (\x\y)--(\x\yi);
    \foreach \x [count=\xi] in {0,...,5}
      \foreach \y in {0,...,5}
	\draw (\x\y)--(\xi\y) ;
    \foreach \x [count=\xi] in {0,...,5}
      \foreach \y [count=\yi] in {0,...,4}
	\draw (\x\y)--(\xi\yi);
    \node (dummy) at (0,-0.15) {};
  \end{tikzpicture}}
\caption{\textbf{Grid Graph.} Regular square tiling (a) and three example
  neighborhoods (b--d) for a $2$-dimensional grid graph derived from line graphs $L_7$ and~$L_6$.}
\label{fig:grid_graph}
\end{figure}

\paragraph{Basic Definitions\\}
Given that the neighborhood of a node is the subgraph induced by all
its adjacent vertices, we define a $d$-dimensional \emph{grid graph}
as a lattice graph whose node embedding in $\bb R^d$ forms a regular
square tiling and the neighborhoods $\cm N$ of each non-border node
are isomorphic (ignoring the node information).
Figure~\ref{fig:grid_graph} illustrates a regular square tiling and
several isomorphic neighborhoods of a $2$-dimensional grid graph.  If
we ignore boundary nodes, a grid graph is a regular graph; i.e., each
non-border node has the same degree. Note that the size of the border
depends on the radius of the neighborhood.  In order to be able to
neglect the special treatment for border nodes, it is common to view
the actual grid graph as a finite section of an actually infinite
graph.

A grid graph whose node embedding in $\bb R^d$ forms a regular square
tiling can be derived from the graph Cartesian product of line
graphs. So, a two-dimensional grid is defined as
\begin{equation}
  G^{(i)} = L_{m_{i,1}} \times L_{m_{i,2}}, \notag
\end{equation}
where $L_{m_{i,1}}$ is a line graph with $m_{i,1}$ nodes. $G^{(i)}$
consists of $n_i = m_{i,1} \, m_{i,2}$ nodes, where non-border nodes
have the same number of neighbors.  Note that the grid graph $G^{(i)}$
only specifies the node layout in the graph but not the edge
structure. The edges are given by the neighborhood $\cm N$ which can
be defined by any arbitrary matrix $B$ encoding the weighted adjacency
of its center node. The nodes, being for instance image pixels, can
carry discrete or continuous vector-valued information.  Thus, in the
most-general setting the database of grid graphs is given by
$\mathbf{G} = \{G^{(i)}\}_{i=1,\ldots,n}$ with $G^{(i)} = (V^{(i)},
\cm N, \ell)$, where $\ell: V^{(i)} \rightarrow \cm L$ with $\cm L =
([k], \bb R^D)$.  Commonly used neighborhoods $\cm N$ are the
4-neighborhood and the $8$-neighborhood illustrated in
Figure~\ref{fig:grid_graph}(b) and (c).

\paragraph{Discrete Convolution\\}
The general convolution operation on two functions $f$ and $g$ is
defined as
\begin{align}
  f(x) \ast g(x) = (f \ast g) (x) = \int_{-\infty}^{\infty} f(\tau) \, g(x-\tau) \,\mathrm{d}\tau. \notag
\end{align}
That is, the convolution operation produces a modified, or
\emph{filtered}, version of the original function $f$.  The function
$g$ is called a filter. For two-dimensional grid graphs interpreted as
discrete functions of two variables $x$ and $y$, e.g., the pixel
location, we consider the discrete spatial convolution defined by:
\begin{align}
  f(x,y) \ast g(x,y) = (f \ast g) (x,y) = \sum_{i = -\infty}^{\infty} \sum_{j = -\infty}^{\infty} f(i,j) \, g(x-i, y-j), \notag
\end{align}
where the computation is in fact done for finite intervals. As
convolution is a well-studied operation in low-level signal processing
and discrete convolution is a standard operation in digital image
processing, we can resort to highly developed algorithms for its
computation; see for example Chapter 2 in \citep{Jahne05}. Convolutions
can be computed efficiently via the fast Fourier transformation in
$\cm{O}(n_i\,\log\,n_i)$ per graph.

\begin{algorithm}[t]
  \caption{Propagation kernel for grid graphs.}
  \begin{algorithmic}
    \State \textbf{given:} graph database $G = (V,E,\ell)$, $\#$ iterations $t_{\textsc{max}}$, \textcolor{ForestGreen}{filter matrix} $\color{ForestGreen} B$, bin width $w$, metric \textsc{m}, base kernel $\langle \cdot, \cdot \rangle$
    \State \textbf{initialization:} $K \gets 0$, $\color{blue} P^{(i)}_0 \gets \delta_{\ell(V_i)}  \; \forall i$
    \For{$t \gets 0\dotsc t_{\textsc{max}}$}
      \State \textsc{calculate-lsh}($\color{blue}\{P^{(i)}_t \}_i$, $w$, \textsc{m}) 	\Comment{bin node information}
      \ForAll{graphs $G^{(i)}$}
	  \State $compute\;\;  \Phi_{i \cdot} = \phi(G^{(i)}_t)$  \Comment{count bin strengths}
      \EndFor
      \textcolor{blue}{
      \ForAll{graphs $G^{(i)}$ and labels $j$}
	  \State $\color{blue} P^{(i,j)}_{t+1} \gets P^{(i,j)}_{t} \ast B$	\Comment{\textcolor{blue}{discrete convolution}}
      \EndFor}
      \State $K \gets K + \langle \Phi, \Phi\rangle $ 	\Comment{compute and add kernel contribution}
    \EndFor
  \end{algorithmic}
  \label{algo:propKernel_grid}
\end{algorithm}

\paragraph{Efficient Propagation Kernel Computation\\}
Now let $\mathbf{G} = \{G^{(i)}\}_i$ be a database of grid graphs. To
simplify notation, however without loss of generality, we assume
two-dimensional grids $G^{(i)} = L_{m_{i,1}} \times L_{m_{i,2}}$.
Unlike in the case of general graphs, each graph now has a natural
two-dimensional structure, so we will update our notation to reflect
this structure. Instead of representing the label probability
distributions of each node as rows in a two-dimensional matrix, we now
represent them in the third dimension of a three-dimensional tensor
$P^{(i)}_t \in \bb{R}^{m_{i,1}\times m_{i,2} \times k}$.  Modifying
the structure makes both the exposition more clear and also enables
efficient computation.  Now, we can simply consider discrete
convolution on $k$ matrices of label probabilities $P^{(i,j)}$ per
grid graph $G^{(i)}$, where $P^{(i,j)} \in \bb{R}^{m_{i,1}\times
  m_{i,2}}$ contains the probabilities of all nodes in $G^{(i)}$ of
being label $j$ and $j \in \{1, \dots, k\}$. For observed labels,
$P^{(i)}_0$ is again initialized with a Kronecker delta distribution
across the third dimension and, in each propagation step, we perform a
discrete convolution of each matrix $P^{(i,j)}$ per graph.  Thus, we
can create various propagation schemes efficiently by applying
appropriate filters, which are represented by matrices $B$ in our
discrete case.  We use circular symmetric neighbor sets $N_{r,p}$ as
introduced in \citep{OjalaPM02}, where each pixel has $p$ neighbors
which are equally spaced pixels on a circle of radius $r$. We use the
following approximated filter matrices in our experiments:
\begin{align}
N_{1,4} &=
  \begin{bmatrix}
    0 & 0.25 & 0\\
    0.25 & 0 & 0.25\\
    0 & 0.25 & 0\\
  \end{bmatrix},\;
N_{1,8} =
  \begin{bmatrix}
    0.06 & 0.17 & 0.06\\
    0.17 & 0.05 & 0.17\\
    0.06 & 0.17 & 0.06\\
  \end{bmatrix},\; \text{and } \notag \\
N_{2,16} &=
  \begin{bmatrix}
    0.01&    0.06&    0.09&    0.06&    0.01\\
    0.06&    0.04&    0   &    0.04&    0.06\\
    0.09&    0   &    0   &    0   &    0.09\\
    0.06&    0.04&    0   &    0.04&    0.06\\
    0.01&    0.06&    0.09&    0.06&    0.01\\
  \end{bmatrix}.
  \label{equ:circ_neigh}
\end{align}
The propagation kernel computation for grid graphs is summarized in
Algorithm~\ref{algo:propKernel_grid}, where the specific parts
compared to the general propagation kernel computation
(Algorithm~\ref{algo:propKernel}) are highlighted in green (input) and
blue (computation). Using fast Fourier transformation, the time
complexity of Algorithm~\ref{algo:propKernel_grid} is $\cm{O}(
(t_{\textsc{max}}-1) N \log N + t_{\textsc{max}}\,n^2\,n^{\star})$.
Note that for the purpose of efficient computation,
\textsc{calculate-lsh} has to be adapted to take the label
distributions $\{P^{(i)}_t\}_i$ as a set of 3-dimensional tensors. By
virtue of the invariance of the convolutions used, propagation kernels
for grid graphs are translation invariant, and when using the circular
symmetric neighbor sets they are also 90-degree rotation
invariant. These properties make them attractive for image-based
texture classification. The use of other filters implementing for
instance anisotropic diffusion depending on the local node information
is a straightforward extension.

\section{Experimental Evaluation}
Our intent here is to investigate the power of propagation kernels
(\textsc{pk}s) for graph classification.  Specifically, we ask: \\
\textbf{(Q1)} How sensitive are propagation kernels with respect
              to their parameters, and how should propagation
              kernels be used for graph classification?\\
\textbf{(Q2)} How sensitive are propagation kernels to missing
              and noisy information? \\
\textbf{(Q3)} Are propagation kernels more flexible than state-of-the-art
              graph kernels?\\
\textbf{(Q4)} Can propagation kernels be computed faster than
              state-of-the-art graph kernels while achieving comparable
              classification performance? \\
Towards answering these questions, we consider several evaluation
scenarios on diverse graph datasets including chemical compounds,
semantic image scenes, pixel texture images, and 3\textsc{d} point
clouds to illustrate the flexibility of \textsc{pk}s.

\subsection{Datasets}
The datasets used for evaluating propagation kernels come from a
variety of different domains and thus have diverse properties. We
distinguish graph databases of labeled and attributed graphs, where
attributed graphs usually also have label information on the nodes.
Also, we separate image datasets where we use the pixel grid graphs
from general graphs, which have varying node
degrees. Table~\ref{tab:datasets} summarizes the properties of all
datasets used in our experiments.

\begin{table}[!ht]
  \centering
  \caption{Dataset statistics and properties.}
  \vspace*{1ex}
  \begin{tabular}{lccccccc}
    \toprule
    & \multicolumn{7}{c}{properties} \\
    \cmidrule(l){2-8}
    \multicolumn{1}{c}{dataset} 	& \multirow{2}{*}{\# graphs} 	&median 	& max \# 	& total \#  	& \# node 	& \# graph 	& attr 	\\
    &  				&\# nodes	& nodes		& nodes 	&  labels	& labels	& dim	\\
    \midrule
    \textsc{mutag}		& 188				&17.5		& 28		& 3\,371	& 7		& 2		& -- \\
    \textsc{nci1}			& 4\,110			&27		& 111		& 122\,747	& 37		& 2		& -- \\
    \textsc{nci109}		& 4\,127			&26		& 111		& 122\,494	& 38		& 2		& -- \\
    \textsc{d\&d}			& 1\,178			&241		& 5\,748	& 334\,925	& 82		& 2		& -- \\
    \midrule
    \textsc{msrc9} 		& 221				&40		& 55		& 8\,968	& 10		& 8		& -- \\
    \textsc{msrc21} 		& 563				&76		& 141		& 43\,644	& 24		& 20		& -- \\
    \midrule
    \textsc{db} 			& 41				&964		& 5\,037	& 56\,468	& 5		& 11		& 1  \\
    \textsc{synthetic}		& 300				&100		& 100		& 30\,000	& --		& 2		& 1  \\
    \textsc{enzymes}		& 600				&32		& 126		& 19\,580	& 3		& 6		& 18 \\
    \textsc{proteins}		& 1\,113			&26		& 620		& 43\,471	& 3		& 2		& 1   \\
    \textsc{pro-full}		& 1\,113			&26		& 620		& 43\,471	& 3		& 2		& 29   \\
    \textsc{bzr}			& 405				&35		& 57		& 14\,479	& 10		& 2		& 3   \\
    \textsc{cox2}			& 467				&41		& 56		& 19\,252	& 8		& 2		& 3   \\
    \textsc{dhfr}			& 756				&42		& 71		& 32\,075	& 9		& 2		& 3  \\
    \midrule
    \textsc{brodatz}		& 2\,048			&4\,096		& 4\,096	& 8\,388\,608 	& 3		& 32		& --  \\
    \textsc{plants}		& 2\,957			&4\,725		& 5\,625 	& 13\,587\,375	& 5		& 6		& -- \\
    \bottomrule
  \end{tabular}
  \label{tab:datasets}
\end{table}

\paragraph{Labeled Graphs\\}
For labeled graphs, we consider the following benchmark datasets from
bioinformatics: \textsc{mutag}, \textsc{nci1}, \textsc{nci109}, and
\textsc{d\&d}. \textsc{mutag} contains $188$ sets of mutagenic
aromatic and heteroaromatic nitro compounds, and the label refers to
their mutagenic effect on the Gram-negative bacterium \emph{Salmonella
  typhimurium}~\citep{Debnath91}.  \textsc{nci1} and \textsc{nci109}
are anti-cancer screens, in particular for cell lung cancer and
ovarian cancer cell lines, respectively~\citep{WaleK06}.  \textsc{d\&d}
consists of $1\,178$ protein structures~\citep{DobsonD03}, where the
nodes in each graph represent amino acids and two nodes are connected
by an edge if they are less than $6$ \AA ngstroms apart. The graph
classes are \emph{enzymes} and \emph{non-enzymes}.

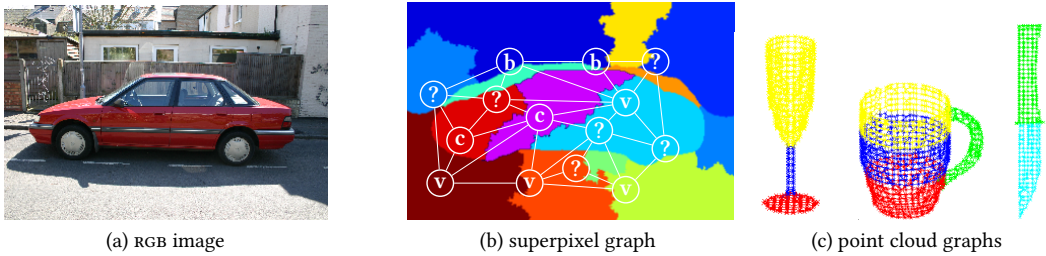
\begin{figure}[t]
  \begin{center}
    \subfloat[\textsc{rgb} image]{\includegraphics[width=0.3\textwidth]{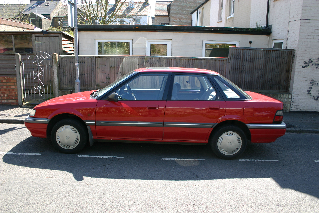}}
    \hfill
    \subfloat[superpixel graph]{\resizebox{0.31\textwidth}{!}{
        \tikzstyle{varnode}=[draw,very thick,circle,white,inner sep=0.05cm,minimum size=8mm]
        \begin{tikzpicture}
          \node[anchor=south west,inner sep=0] at (-0.1,-0.1) {\includegraphics[width=0.699\textwidth]{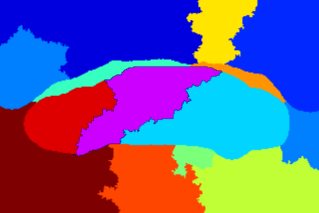}};
          \huge
          \node[varnode] at (3,4.7) (1) {\textbf{b}};
          \node[varnode] at (5.6,4.7) (2) {\textbf{b}};
          \node[varnode] at (7.4,4.7) (3) {\textbf{?}};
          \node[varnode] at (0.7,3.7) (4) {\textbf{?}};
          \node[varnode] at (2.6,3.55) (5) {\textbf{?}};
          \node[varnode] at (3.9,3) (6) {\textbf{c}};
          \node[varnode] at (5.7,2.6) (7) {\textbf{?}};
          \node[varnode] at (6.5,3.45) (8) {\textbf{v}};
          \node[varnode] at (7.7,2.05) (9) {\textbf{?}};
          \node[varnode] at (0.9,1) (10) {\textbf{v}};
          \node[varnode] at (1.5,2.3) (11) {\textbf{c}};
          \node[varnode] at (3.6,1) (12) {\textbf{v}};
          \node[varnode] at (5,1.45) (13) {\textbf{?}};
          \node[varnode] at (6.5,0.8) (14) {\textbf{v}};
          \draw[thick,white] (1) -- (2);
          \draw[thick,white] (1) -- (4);
          \draw[thick,white] (1) -- (5);
          \draw[thick,white] (1) -- (8);
          \draw[thick,white] (2) -- (3);
          \draw[thick,white] (2) -- (8);
          \draw[thick,white] (3) -- (8);
          \draw[thick,white] (3) -- (9);
          \draw[thick,white] (4) -- (5);
          \draw[thick,white] (4) -- (10);
          \draw[thick,white] (4) -- (11);
          \draw[thick,white] (5) -- (6);
          \draw[thick,white] (5) -- (11);
          \draw[thick,white] (5) -- (8);
          \draw[thick,white] (6) -- (7);
          \draw[thick,white] (6) -- (8);
          \draw[thick,white] (6) -- (11);
          \draw[thick,white] (6) -- (12);
          \draw[thick,white] (7) -- (8);
          \draw[thick,white] (7) -- (9);
          \draw[thick,white] (7) -- (12);
          \draw[thick,white] (7) -- (13);
          \draw[thick,white] (7) -- (14);
          \draw[thick,white] (8) -- (9);
          \draw[thick,white] (9) -- (14);
          \draw[thick,white] (10) -- (11);
          \draw[thick,white] (10) -- (12);
          \draw[thick,white] (10) -- (6);
          \draw[thick,white] (12) -- (13);
          \draw[thick,white] (12) -- (14);
          \draw[thick,white] (13) -- (14);
      \end{tikzpicture}}}
    \subfloat[point cloud graphs]{\includegraphics[scale=0.29]{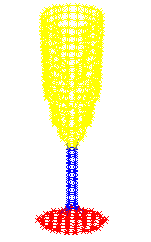}
				  \includegraphics[scale=0.21]{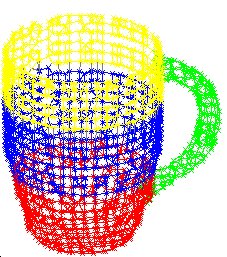}
				  \includegraphics[scale=0.21]{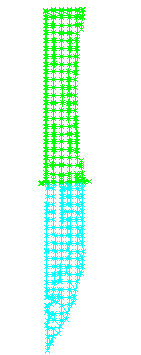}
				  \label{fig:obj_graphs}}
    \caption[]{\textbf{Semantic Scene and Point Cloud Graphs.} The
      \textsc{rgb} image in (a) is represented by a graph of
      superpixels (b) with semantic labels \textbf{b} $=$
      \textit{building}, \textbf{c} $=$ \textit{car}, \textbf{v} $=$
      \textit{void}, and \textsc{\textbf{?}} $=$ unlabeled.  (c) shows
      point clouds of household objects represented by labeled
      $4$-\textsc{nn} graphs with part labels \emph{top} (yellow),
      \emph{middle} (blue), \emph{bottom} (red), \emph{usable-area}
      (cyan), and \emph{handle} (green).  Edge colors are derived from
      the adjacent nodes.}
    \label{fig:image_graph}
  \end{center}
\end{figure}

\paragraph{Partially Labeled Graphs\\}
The two real-world image datasets \textsc{msrc} 9-class and
\textsc{msrc}
21-class\footnote{\url{http://research.microsoft.com/en-us/projects/objectclassrecognition/}}
are state-of-the-art datasets in semantic image processing originally
introduced in \citep{WinnCM05}.  Each image is represented by a
conditional Markov random field graph, as illustrated in
Figure~\ref{fig:image_graph}(a) and (b).  The nodes of each graph are
derived by oversegmenting the images using the quick shift
algorithm,\footnote{\url{http://www.vlfeat.org/overview/quickshift.html}}
resulting in one graph among the superpixels of each image. Nodes are
connected if the superpixels are adjacent, and each node can further
be annotated with a semantic label. Imagining an image retrieval
system, where users provide images with semantic information, it is
realistic to assume that this information is only available for parts
of the images, as it is easier for a human annotator to label a small
number of image regions rather than the full image. As the images in
the \textsc{msrc} datasets are fully annotated, we can derive semantic
(ground-truth) node labels by taking the mode ground-truth label of
all pixels in the corresponding superpixel.  Semantic labels are, for
example, \textit{building, grass, tree, cow, sky, sheep, boat, face,
  car, bicycle}, and a label \textit{void} to handle objects that do
not fall into one of these classes.  We removed images consisting of
solely one semantic label, leading to a classification task among
eight classes for \textsc{msrc9} and $20$ classes for \textsc{msrc21}.

\paragraph{Attributed Graphs\\}
To evaluate the ability of \textsc{pk}s to incorporate continuous node
attributes, we consider the attributed graphs used in
\citep{FeragenKPBB13,KriegeM12}. Apart from one synthetic dataset
(\textsc{synthetic}), the graphs are all chemical compounds
(\textsc{enzymes}, \textsc{proteins}, \textsc{pro-full}, \textsc{bzr},
\textsc{cox2}, and \textsc{dhfr}).  \textsc{synthetic} comprises 300
graphs with 100 nodes, each endowed with a one-dimensional normally
distributed attribute and 196 edges each. Each graph class, $A$ and
$B$, has 150 examples, where in $A$, 10 node attributes were flipped
randomly and in $B$, 5 were flipped randomly. Further, noise drawn
from $\cm{N}(0,0.45^2)$ was added to the attributes in $B$.
\textsc{proteins} is a dataset of chemical compounds with two classes
(\emph{enzyme} and \emph{non-enzyme}) introduced
in~\citep{DobsonD03}. \textsc{enzymes} is a dataset of protein tertiary
structures belonging to $600$ enzymes from the \textsc{brenda}
database~\citep{Schomburg04}.  The graph classes are their \textsc{ec}
(enzyme commission) numbers which are based on the chemical reactions
they catalyze.  In both datasets, nodes are secondary structure
elements (\textsc{sse}), which are connected whenever they are
neighbors either in the amino acid sequence or in 3\textsc{d}
space. Node attributes contain physical and chemical measurements
including length of the \textsc{sse} in \r{A}ngstrom, its
hydrophobicity, its van der Waals volume, its polarity, and its
polarizability. For \textsc{bzr}, \textsc{cox2}, and \textsc{dhfr} --
originally used in \citep{maheV09} -- we use the 3\textsc{d}
coordinates of the structures as attributes.

\paragraph{Point Cloud Graphs\\}
In addition, we consider the object database
\textsc{db},\footnote{\url{http://www.first-mm.eu/data.html}}
introduced in \citep{Neumann13mlg}. \textsc{db} is a collection of $41$
simulated 3\textsc{d} point clouds of household objects.  Each object
is represented by a labeled graph where nodes represent points, labels
are semantic parts (\emph{top}, \emph{middle}, \emph{bottom},
\emph{handle}, and \emph{usable-area}), and the graph structure is
given by a $k$-nearest neighbor ($k$-\textsc{nn}) graph
w.r.t.\ Euclidean distance of the points in 3\textsc{d} space,
cf.\ Figure~\ref{fig:image_graph}(c).  We further endowed each node
with a continuous curvature attribute approximated by its derivative,
that is, by the tangent plane orientations of its incident nodes.  The
attribute of node $u$ is given by $x_{u} = \sum_{v \in \cm{N}(u)}
1-|\mathbf{n}_u \cdot \mathbf{n}_v|$, where $\mathbf{n}_u$ is the
normal of point~$u$ and $\cm{N}(u)$ are the neighbors of node $u$.
The classification task here is to predict the category of each
object. Examples of the $11$ categories are \emph{glass}, \emph{cup},
\emph{pot}, \emph{pan}, \emph{bottle}, \emph{knife}, \emph{hammer},
and \emph{screwdriver}.

\paragraph{Grid Graphs\\}
We consider a classical benchmark dataset for texture classification
(\textsc{brodatz}) and a dataset for plant disease classification
(\textsc{plants}).  All graphs in these datasets are grid graphs
derived from pixel images. That is, the nodes are image pixels
connected according to circular symmetric neighbor sets $N_{r,p}$ as
exemplified in Eq.~\eqref{equ:circ_neigh}.  Node labels are computed
from the \textsc{rgb} color values by quantization.

\begin{figure}[t]
\vspace*{-0.2cm}
  \subfloat[bark]{\includegraphics[scale=1.2]{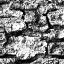}} \hfill
  \subfloat[grass]{\includegraphics[scale=1.2]{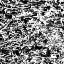}} \hfill
  \subfloat[phoma]{\includegraphics[scale=1.03,angle=90]{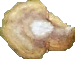}}  \hfill
  \subfloat[cercospora]{\includegraphics[scale=1.03,angle=90]{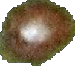}}  \\
  \subfloat[bark-3]{\includegraphics[scale=0.12]{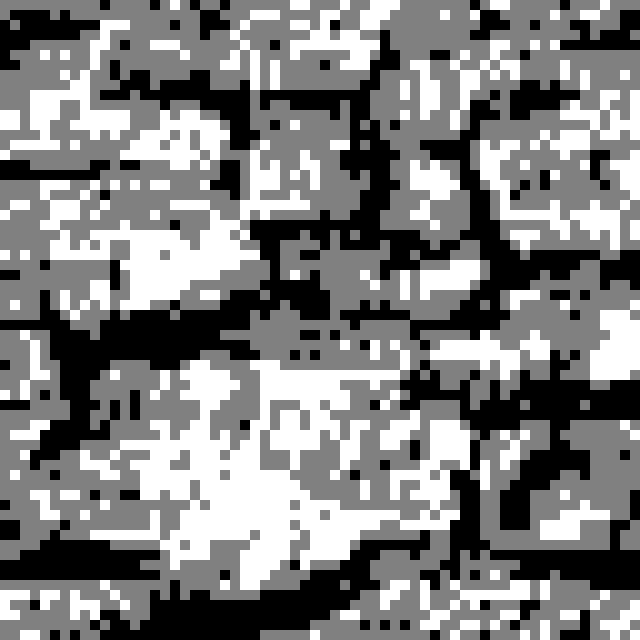}} \hfill
  \subfloat[grass-3]{\includegraphics[scale=0.12]{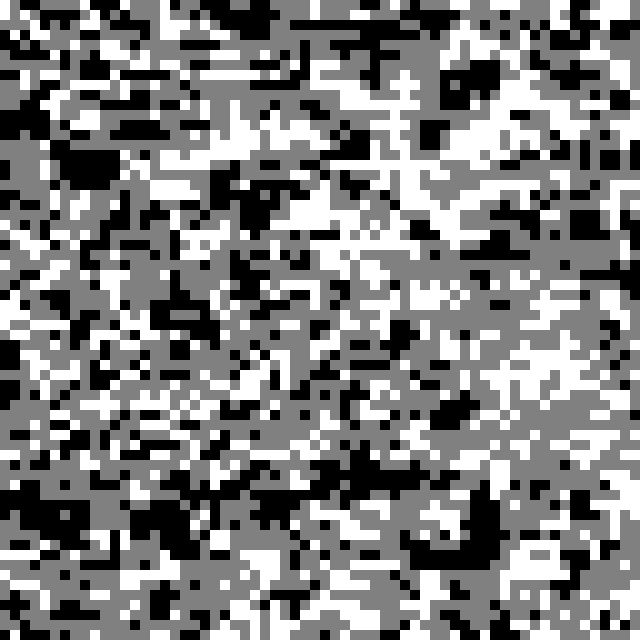}}  \hfill
  \subfloat[phoma-5]{\includegraphics[scale=1.03,angle=90]{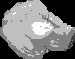}} \hfill
  \subfloat[cercospora-5]{\includegraphics[scale=1.03,angle=90]{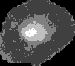}}
  \caption{Example images from \textsc{brodatz} (a,b) and
    \textsc{plants} (c,d) and the corresponding quantized versions
    with 3 colors (e,f) and 5 colors (g,h).}
  \label{fig:plant_regions}
  \vspace*{-0.2cm}
\end{figure}

\textsc{brodatz},\footnote{\url{http://www.ee.oulu.fi/research/imag/texture/image_data/Brodatz32.html}}
introduced in~\citep{ValkealahtiO98}, covers 32 textures from the
Brodatz album with 64 images per class comprising the following
subsets of images: 16 ``original'' images (\textsc{o}), 16 rotated
versions (\textsc{r}), 16 scaled versions (\textsc{s}), and 16 rotated
and scaled versions (\textsc{rs}) of the ``original''
images. Figure~\ref{fig:plant_regions}(a) and (b) show example images
with their corresponding quantized versions (e) and (f).  For
parameter learning, we used a random subset of 20\% of the original
images and their rotated versions, and for evaluation we use test
suites similar to the ones provided with the dataset.\footnote{The
  test suites provided with the data are incorrect.} All
train/test splits are created such that whenever an original image
(\textsc{o}) occurs in one split, their modified versions
(\textsc{r,s,rs}) are also included in the same split.

The images in \textsc{plants}, introduced in \citep{NeumannHKKB14}, are
regions showing disease symptoms extracted from a database of $495$
\textsc{rgb} images of beet leaves. The dataset has six classes: five
disease symptoms \emph{cercospora}, \emph{ramularia},
\emph{pseudomonas}, \emph{rust}, and \emph{phoma}, and one class for
extracted regions not showing a disease symptom.
Figure~\ref{fig:plant_regions}(c) and (d) illustrates two regions and
their quantized versions (g) and (h).  We follow the experimental
protocol in \citep{NeumannHKKB14} and use $10\%$ of the full data
covering a balanced number of classes ($296$ regions) for parameter
learning and the full dataset for evaluation.  Note that this dataset
is highly imbalanced, with two infrequent classes accounting for only
$2\%$ of the examples and two frequent classes covering $35\%$ of the
examples.

\subsection{Experimental Protocol}
We implemented propagation kernels in Matlab\footnote{All
  implementations of \textsc{pk} and \textsc{p2k} are available here:
  \url{https://github.com/marionmari/propagation_kernels}} and
classification performance on all datasets except for \textsc{db} is
evaluated by running \textsc{c-svm} classifications using
\texttt{libSVM}.\footnote{\url{http://www.csie.ntu.edu.tw/~cjlin/libsvm/}}
For the sensitivity analysis, the cost parameter $c$ was set to its
default value of 1 for all datasets, whereas for the experimental
comparison with existing graph kernels, we learned it via 5-fold
cross-validation on the training set for all methods.  The number of
kernel iterations $t_{\textsc{max}}$ was learned on the training
splits.  Reported accuracies are an average of $10$ reruns of a
stratified $10$-fold cross-validation.

For \textsc{db,} we follow the protocol introduced in
\citep{Neumann13mlg}. We perform a leave-one-out (\textsc{loo}) cross
validation on the 41 objects in \textsc{db}, where the kernel
parameter $t_{\textsc{max}}$ is learned on each training set again via
\textsc{loo}.  We further enhanced the nodes by a standardized
continuous curvature attribute, which was previously only encoded in
the edge weights.  For all \textsc{pk}s, the \textsc{lsh} bin-width
parameters were set to $w_l = 10^{-5}$ for labels and to $w_a = 1$ for
the normalized attributes, and as \textsc{lsh} metrics we chose
$\textsc{m}_l = \textsc{tv}$ and $\textsc{m}_a = \textsc{l1}$ in all
experiments.  Before we evaluate classification performance and
runtimes of the proposed propagation kernels, we analyze their
sensitivity towards the choice of kernel parameters and with respect
to missing and noisy observations.

\subsection{Parameter Analysis}
\begin{figure}[t!]
  \begin{center}
  \begin{minipage}{\textwidth}
    \subfloat[\textsc{mutag}]{\includegraphics[width=0.2\textwidth]{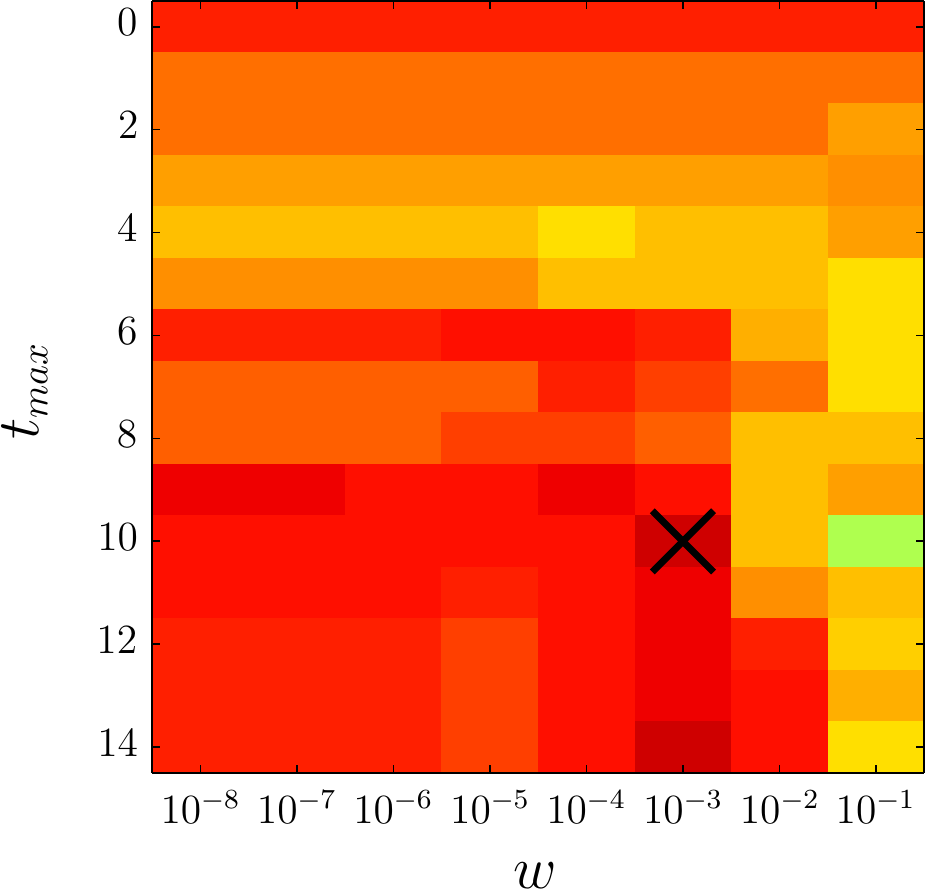}}
    \hfill  \subfloat[\textsc{mutag},normalized][\shortstack{\textsc{mutag}\\normalized}]{\includegraphics[width=0.2\textwidth]{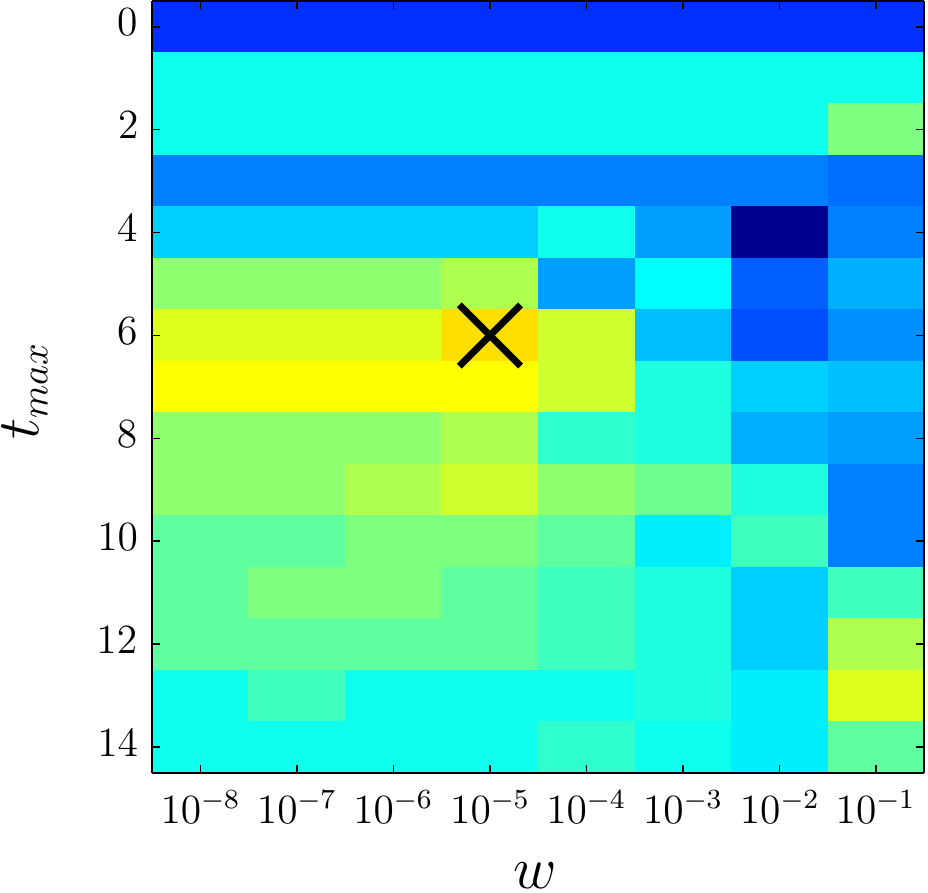}}
    \hfill  \includegraphics[width=0.04\textwidth]{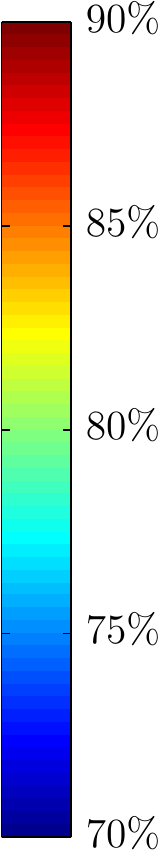} \hspace{0.3cm}
    \subfloat[\textsc{enzymes}]{\includegraphics[width=0.2\textwidth]{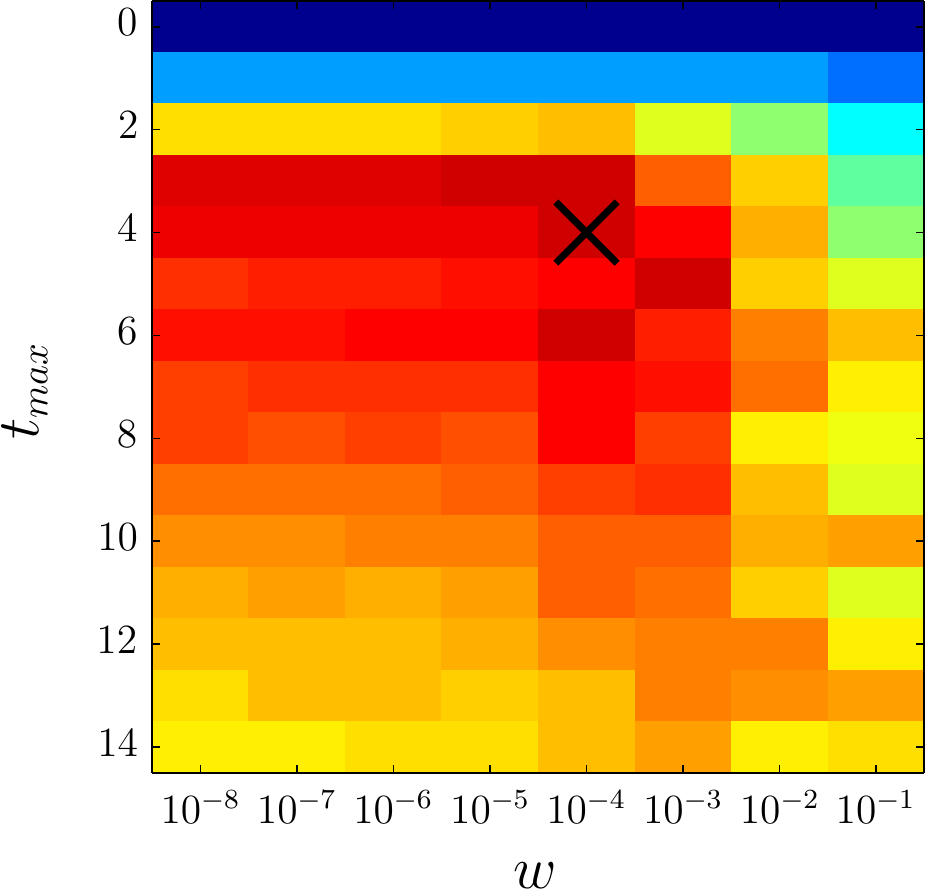}}
    \hfill  \subfloat[\textsc{enzymes}, normalized][\shortstack{\textsc{enzymes}\\normalized}]{\includegraphics[width=0.2\textwidth]{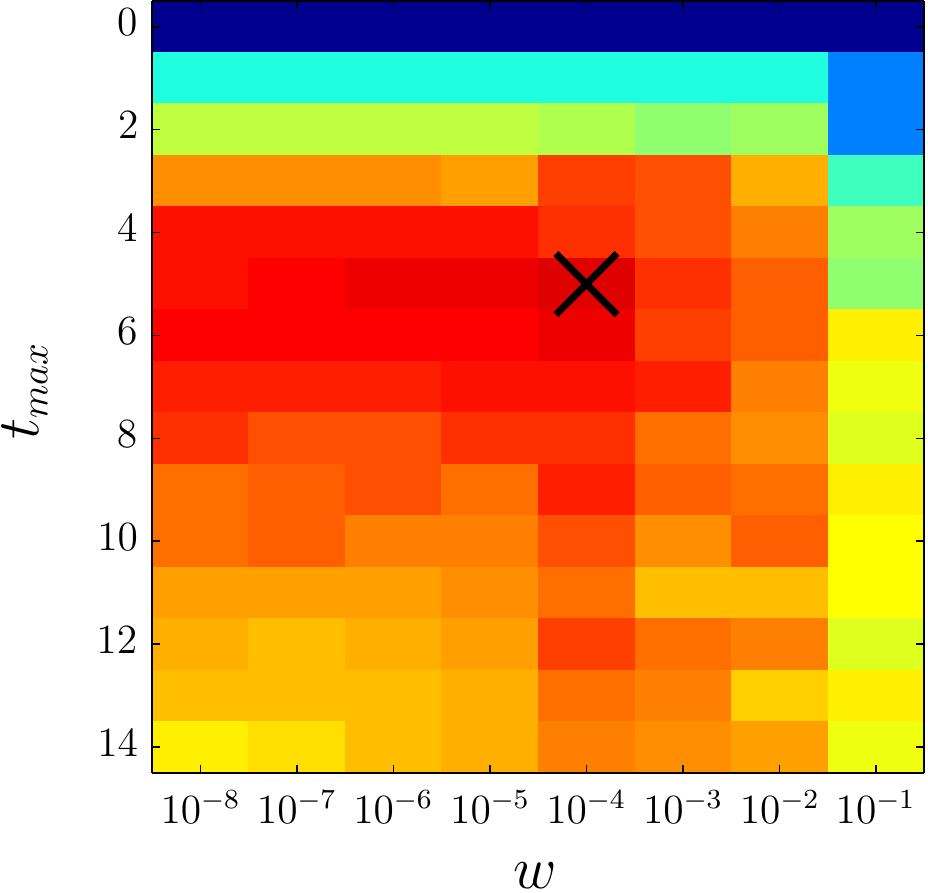}}
    \hfill  \includegraphics[width=0.04\textwidth]{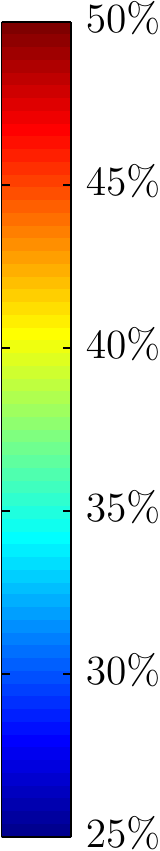} \\
    \subfloat[\textsc{nci1}]{\includegraphics[width=0.2\textwidth]{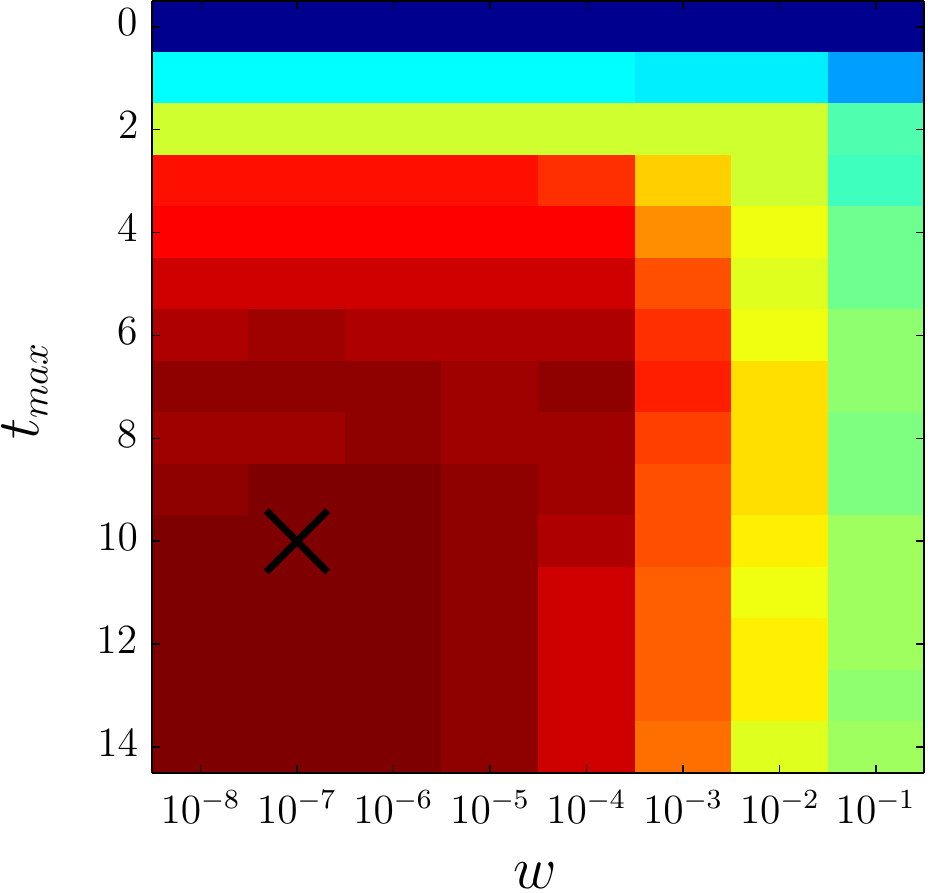}}
    \hfill  \subfloat[\textsc{nci1},normalized][\shortstack{\textsc{nci1}\\normalized}]{\includegraphics[width=0.2\textwidth]{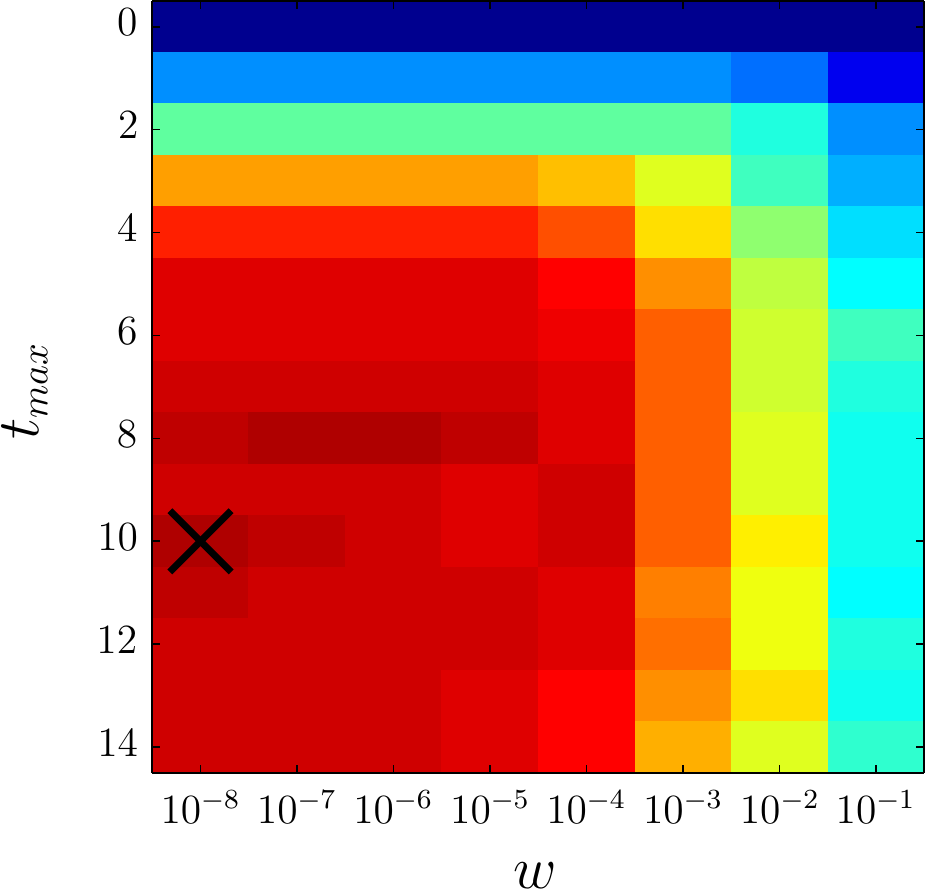}}
    \hfill  \includegraphics[width=0.04\textwidth]{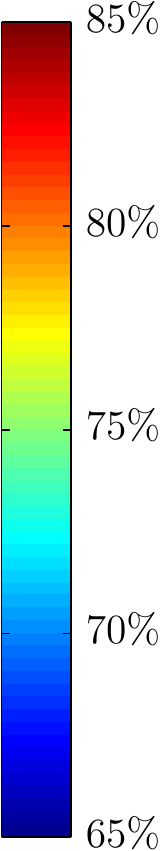}\hspace{0.3cm}
    \subfloat[\textsc{db}]{\includegraphics[width=0.2\textwidth]{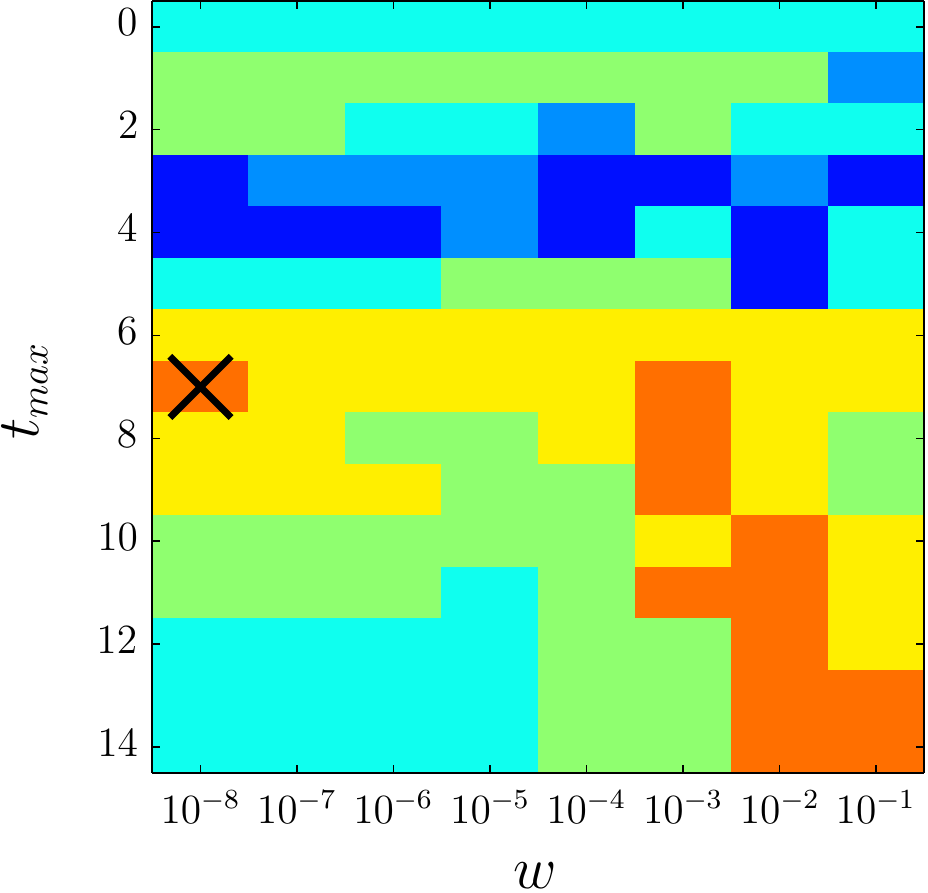}}
    \hfill  \subfloat[\textsc{db},normalized][\shortstack{\textsc{db}\\normalized}]{\includegraphics[width=0.2\textwidth]{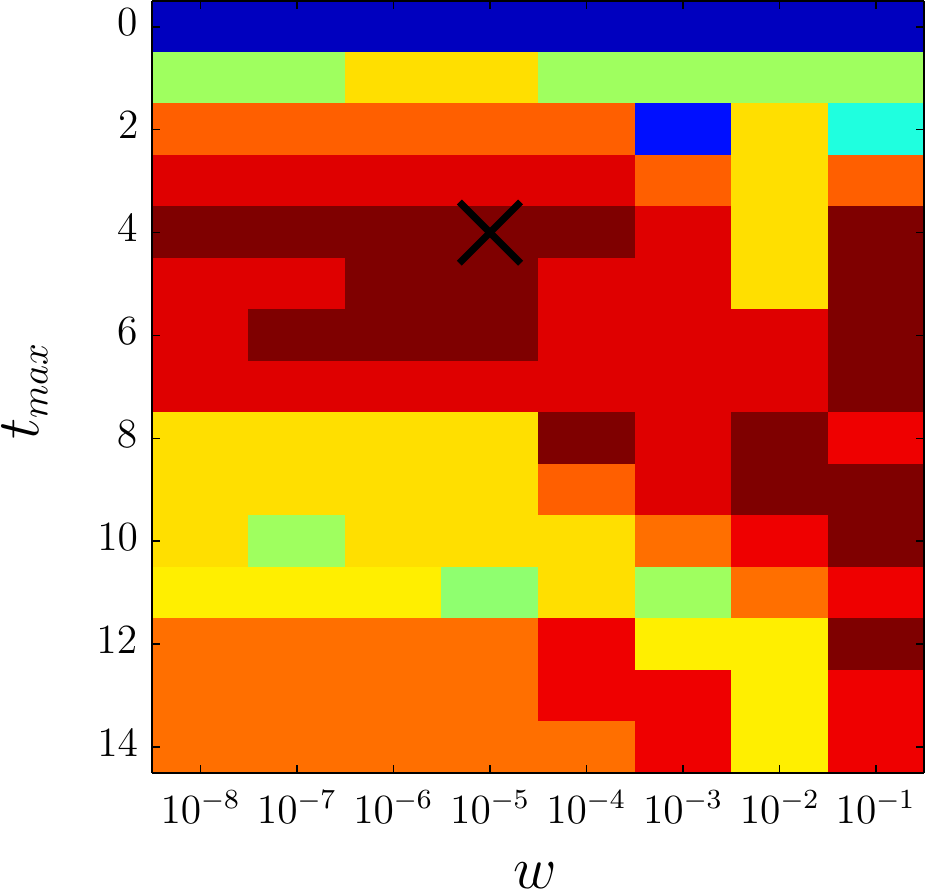}}
    \hfill  \includegraphics[width=0.04\textwidth]{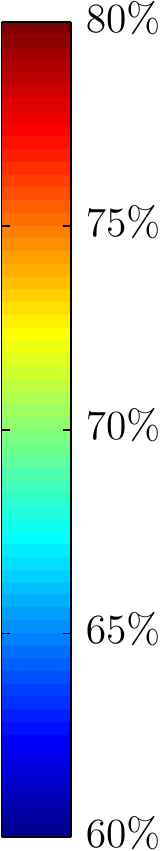}
  \end{minipage}
  \caption[]{\textbf{Parameter Sensitivity of \textsc{pk}.} The plots
    show heatmaps of average accuracies (10-fold \textsc{cv}) of
    \textsc{pk} (labels only) w.r.t.\ the bin widths parameter $w$ and
    the number of kernel iterations $t_{\textsc{max}}$ for four
    datasets \textsc{mutag}, \textsc{enzymes}, \textsc{nci1}, and
    \textsc{db}.  In panels (a, c, e, g) we used the kernel matrix
    directly, in panels (b, d, f, h) we normalized the kernel matrix.
    The \textsc{svm} cost parameter is learned for each combination of
    $w$ and $t_{\textsc{max}}$ on the full
    dataset. $\boldsymbol{\times}$ marks the highest accuracy.}
  \label{fig:heatmaps}
  \end{center}
\end{figure}

To analyze parameter sensitivity with respect to the kernel parameters
$w$ (\textsc{lsh} bin width) and $t_{\textsc{max}}$ (number of kernel
iterations), we computed average accuracies over 10 randomly generated
test sets for all combinations of $w$ and $t_{\textsc{max}}$, where $w
\in \{10^{-8},10^{-7}, \dots, 10^{-1} \}$ and $t_{\textsc{max}} \in
\{0,1,\dots,14\}$ on \textsc{mutag}, \textsc{enzymes}, \textsc{nci1},
and \textsc{db}.  The propagation kernel computation is as described
in Algorithm~\ref{algo:propKernel_labeled}, that is, we used the label
information on the nodes and the label diffusion process as
propagation scheme.  To assess classification performance, we first
learn the \textsc{svm} cost parameter on the full dataset for each
parameter combination and then performed a 10-fold cross validation
(\textsc{cv}). Further, we repeated each of these experiments with the
normalized kernel, where normalization means dividing each kernel
value by the square root of the product of the respective diagonal
entries. Note that for normalized kernels we test for larger
\textsc{svm} cost values. Figure~\ref{fig:heatmaps} shows heatmaps of
the results.

In general, we see that the \textsc{pk} performance is relatively
smooth, especially if $w < 10^{-3}$ and $t_{\textsc{max}} >
4$. Specifically, the number of iterations leading to the best results
are in the range from $\{4,\dots,10\}$ meaning that we do not have to
use a larger number of iterations in the \textsc{pk} computations,
helping to keep a low computation time. This is especially important
for parameter learning.  Comparing the heatmaps of the normalized
\textsc{pk} to the unnormalized \textsc{pk} leads to the conclusion
that normalizing the kernel matrix can actually hurt performance.  For
\textsc{mutag}, Figure~\ref{fig:heatmaps}(a)
and~\ref{fig:heatmaps}(b), the performance drops from $88.2\%$ to
$82.9\%$, indicating that for this dataset the size of the graphs, or
more specifically the amount of labels from the different kind of node
classes, are a strong class indicator for the graph
label. Nevertheless, incorporating the graph structure, i.e.,
comparing $t_{\textsc{max}}=0$ to $t_{\textsc{max}}=10$, can still
improve classification performance by $1.5\%$. For other prediction
scenarios such as the object category prediction on the \textsc{db}
dataset, Figure~\ref{fig:heatmaps}(g) and~\ref{fig:heatmaps}(h), we
actually want to normalize the kernel matrix to make the prediction
independent of the object scale. That is, a cup scanned from a larger
distance being represented by a smaller graph is still a cup and
should be similar to a larger cup scanned from a closer view. So, for
our experiments on object category prediction we will use normalized
graph kernels whereas for the chemical compounds we will use
unnormalized kernels unless stated otherwise.

In summary, we can answer {\bf (Q1)} by concluding that \textsc{pk}s
are not overly sensitive to the choice of parameters and we propose to
learn $t_{\textsc{max}} \in \{0,1,\dots,10\}$ and fix $w \leq
10^{-3}$.  Further, we recommend to decide on using the normalized
version of \textsc{pk}s only when graph size invariance is deemed
important for the classification task.

\subsection{Comparison to Existing Graph Kernels}
We compare classification accuracy and runtime of propagation kernels
(\textsc{pk}) with the following state-of-the-art graph kernels: the
Weisfeiler--Lehman subtree kernel (\textsc{wl})
\citep{ShervashidzeB11}, the shortest path kernel (\textsc{sp})
\citep{borgwardtK05}, the graph hopper kernel (\textsc{gh})
\citep{FeragenKPBB13}, and the common subgraph matching kernel
(\textsc{csm}) \citep{KriegeM12}.  Table~\ref{tab:kernels} lists all
graph kernels and the types of information they are intended for.  For
all \textsc{wl} computations, we used the fast
implementation\footnote{\url{https://github.com/rmgarnett/fast_wl}}
introduced in \citep{KerstingMGG14}. In \textsc{sp}, \textsc{gh}, and
\textsc{csm}, we used a Dirac kernel to compare node labels and a
Gaussian kernel $k_a(u,v) = \exp(-\gamma \| x_u - x_v\|^2)$ with
$\gamma = \nicefrac{1}{D}$ for attribute information, if
feasible. \textsc{csm} for the bigger datasets (\textsc{enzymes},
\textsc{proteins}, \textsc{synthetic}) was computed using a Gaussian
truncated for inputs with $\|x_u - x_v\| > 1$.  We made this decision
to encourage sparsity in the generated (node) kernel matrices,
reducing the size of the induced product graphs and speeding up
computation. Note that this is technically not a valid kernel between
nodes; nonetheless, the resulting graph kernels were always positive
definite. For \textsc{pk} and \textsc{wl} the number of kernel
iterations ($t_{\textsc{max}}$ or\ $h_{\textsc{max}}$) and for
\textsc{csm} the maximum size of subgraphs ($k$) was learned on the
training splits via 10-fold cross validation. For all runtime
experiments all kernels were computed for the largest value of
$t_{\textsc{max}}$, $h_{\textsc{max}}$, or $k$, respectively.  We used
a linear base kernel for all kernels involving count features, and
attributes, if present, were standardized.  Further, we considered
several baselines that do not take the graph structure into
account. \textsc{labels}, corresponding to a \textsc{pk} with
$t_{\textsc{max}}=0$, only compares the label proportions in the
graphs, \textsc{a} takes the mean of a Gaussian node kernel among all
pairs of nodes in the respective graphs, and \textsc{a lsh} again
corresponds to a \textsc{pk} with $t_{\textsc{max}}=0$ using the
attribute information only.
\begin{table}[t]
  \centering
  \caption{Graph kernels and their intended use.}
  \vspace*{1ex}
  \begin{tabular}{c cccccc}
    \toprule
    &\multicolumn{6}{c}{information type} \\
    \cmidrule(l){2-7}
    \multirow{2}{*}{kernel} &\textsc{node} 		& \textsc{partial} & \textsc{edge}    & \textsc{edge}   &\textsc{node} & \textsc{grid graphs}	\\
    &\textsc{labels} 	& \textsc{labels}  & \textsc{weights} & \textsc{labels}	&\textsc{attributes} & \textsc{(fast scaling)} \\
    \midrule
    \textsc{pk}  	& yes & yes & yes & --  & yes & yes \\
    \textsc{wl}     & yes & --  & --  & --  & --  & -- \\
    \textsc{sp}     & yes & --  & yes & yes & yes & -- \\
    \textsc{gh}     & yes & --  & yes & --  & yes & -- \\
    \textsc{csm}    & yes & --  & yes & yes & yes & -- \\
    \bottomrule
  \end{tabular}
  \label{tab:kernels}
\end{table}

\subsubsection{Graph Classification on Benchmark Data}
In this section, we consider graph classification for fully labeled,
partially labeled, and attributed graphs.

\paragraph{Fully Labeled Graphs\\}
The experimental results for labeled graphs are shown in
Table~\ref{tab:acc_labels}.  On \textsc{mutag}, the baseline using
label information only (\textsc{labels}) already gives the best
performance indicating that for this dataset the actual graph
structure is not adding any predictive information.  On \textsc{nci1}
and \textsc{nci109}, \textsc{wl} performs best; however, propagation
kernels come in second while being computed over one minute
faster. Although \textsc{sp} can be computed quickly, it performs
significantly worse than \textsc{pk} and \textsc{wl}. The same holds
for \textsc{gh}, where for this kernel the computation is
significantly slower.  In general, the results on labeled graphs show
that propagation kernels can be computed faster than state-of-the-art
graph kernels but achieve comparable classification performance, thus
question {\bf (Q4)} can be answered affirmatively.

\begin{table}[!t]
  \centering
  \caption{\textbf{Labeled Graphs.} Average accuracies $\pm$ standard
    error of 10-fold cross validation (10 runs).  Average runtimes in
    sec ($x''$), min ($x'$), or hours ($xh$) are given in parentheses.
    The kernel parameters $t_{\textsc{max}}$ for \textsc{pk} and
    $h_{\textsc{max}}$ for \textsc{wl} were learned on the training
    splits ($t_{\textsc{max}}, h_{\textsc{max}} \in
    \{0,1,\dots,10\}$).  \textsc{labels} corresponds to \textsc{pk}
    with $t_{\textsc{max}}=0$.  \scriptsize OUT OF TIME \normalsize
    indicates that the kernel computation did not finish within $24$h.
    Bold indicates that the method performs significantly better than
    the second best method under a paired $t$-test ($p<0.05$).}
  \vspace*{1ex}
  \begin{tabular}{p{1.0cm}cccc}
    \toprule
    & \multicolumn{4}{c}{dataset} \\
    \cmidrule(l){2-5}
    method 	 		&\textsc{mutag} 		& \textsc{nci1} 		& \textsc{nci109} 		& \textsc{dd}	\\
    \midrule
    \textsc{pk}   		&$84.5$ $\pm 0.6$ $(0.2'')$&$84.5$ $\pm 0.1$ $(4.5'$)&$83.5$ $\pm 0.1$ $(4.4'$)&$78.8$ $\pm 0.2$ $(3.6'$)\\
    \textsc{wl}   		&$84.0$ $\pm 0.4$ $(0.2'')$&$\mathbf{85.9}$ $\pm 0.1$ $(5.6'$)&$\mathbf{85.9}$ $\pm 0.1$ $(7.4'$)&$79.0$ $\pm 0.2$ $(6.7'$)\\
    \textsc{sp}   		&$85.8$ $\pm 0.2$ $(0.2'')$	&$74.4$ $\pm 0.1$ $(21.3'')$	&$73.7$ $\pm 0.0$  $(19.3'')$ 	&\tiny OUT OF TIME \\
    \textsc{gh}   		&$85.4$ $\pm 0.5$ $(1.0'$)	&$73.2$ $\pm 0.1$ $(13.0$h)	& $72.6$ $\pm 0.1$ $(22.1$h)  & $68.9$ $\pm 0.2$ ($69.1$h)\\
    \midrule
    \textsc{labels}  	&$85.8$ $\pm 0.2$ $(0.2'')$&$64.6$ $\pm 0.0$ $(4.5'$)&$63.6$ $\pm 0.0$ $(4.4'$)&$78.4$ $\pm 0.1$ $(3.6'$)\\
    \bottomrule
  \end{tabular}
  \label{tab:acc_labels}
\end{table}

\paragraph{Partially Labeled Graphs\\}
To assess the predictive performance of propagation kernels on
partially labeled graphs, we ran the following experiments $10$ times.
We randomly removed 20--80\% of the node labels in all graphs in
\textsc{msrc9} and \textsc{msrc21} and computed cross-validation
accuracies and standard errors. Because the \textsc{wl}-subtree kernel
was not designed for partially labeled graphs, we compare \textsc{pk}
to two variants: one where we treat unlabeled nodes as an additional
label ``$u$'' (\textsc{wl}) and another where we use hard labels
derived from running label propagation (\textsc{lp}) until convergence
(\textsc{lp} $+$ \textsc{wl}).  The results are shown in
Table~\ref{tab:acc_results_partial}.  For larger fractions of missing
labels, \textsc{pk} obviously outperforms the baseline methods, and,
surprisingly, running label propagation until convergence and then
computing \textsc{wl} gives slightly worse results than \textsc{wl}.
However, label propagation might be beneficial for larger amounts of
missing labels.  The runtimes of the different methods on
\textsc{msrc21} are shown in Figure~\ref{fig:image_run} in the
Appendix.  \textsc{wl} computed via the string-based implementation
suggested in \citep{ShervashidzeB11} is over $36$ times slower than
\textsc{pk}. These results again confirm that propagation kernels have
attractive scalability properties for large datasets. The \textsc{lp}
$+$ \textsc{wl} approach wastes computation time while running
\textsc{lp} to convergence before it can even begin calculating the
kernel. The intermediate label distributions obtained during the
convergence process are already extremely powerful for classification.
These results clearly show that propagation kernels can successfully
deal with partially labeled graphs and suggest an affirmative answer
to questions {\bf (Q3)} and {\bf (Q4)}.

\begin{table}[!t]
  \centering
  \caption{\textbf{Partially Labeled Graphs.} Average accuracies (and
    standard errors) on $10$ different sets of partially labeled images
    for \textsc{pk} and \textsc{wl} with unlabeled nodes treated as
    additional label (\textsc{wl}) and with hard labels derived from
    converged label propagation (\textsc{lp} $+$ \textsc{wl}).  Bold
    indicates that the method performs significantly better than the
    second best method under a paired $t$-test ($p<0.05$).}
  \vspace*{1ex}
  \begin{tabular}{llcccc}
    \toprule
    & & \multicolumn{4}{c}{labels missing} \\
    \cmidrule(l){3-6}
    dataset &
    \multicolumn{1}{c}{method} &
    $20\%$ &
    $40\%$ &
    $60\%$ &
    $80\%$ \\
    \midrule
    & \textsc{pk} &
    $90.0$ $\pm0.4$ &
    $88.7$ $\pm0.3$ &
    $86.6$ $\pm0.4$ &
    $80.4$ $\pm0.6$ \\
    \textsc{msrc9}\hspace{1em} & \textsc{lp} $+$ \textsc{wl} &
    $90.0$ $\pm0.2$ &
    $87.9$ $\pm0.6$ &
    $83.2$ $\pm0.6$ &
    $77.9$ $\pm1.0$ \\
    & \textsc{wl} &
    $89.2$ $\pm0.5$ &
    $88.1$ $\pm0.5$ &
    $85.7$ $\pm0.6$ &
    $78.5$ $\pm0.9$ \\
    \midrule
    & \textsc{pk} &
    $\mathbf{86.9}$ $\pm0.3$ &
    $\mathbf{84.7}$ $\pm0.3$ &
    $\mathbf{79.5}$ $\pm0.3$ &
    $\mathbf{69.3}$ $\pm0.3$ \\
    \textsc{msrc21}\hspace{1em}
    &  \textsc{lp} $+$ \textsc{wl}  &
    $85.8$ $\pm0.2$ &
    $81.5$ $\pm0.3$ &
    $74.5$ $\pm0.3$ &
    $64.0$ $\pm0.4$ \\
    & \textsc{wl} &
    $85.4$ $\pm0.4$ &
    $81.9$ $\pm0.4$ &
    $76.0$ $\pm0.3$ &
    $63.7$ $\pm0.4$ \\
    \bottomrule
  \end{tabular}
  \label{tab:acc_results_partial}
\end{table}

\begin{figure}[t!]
  \begin{center}
  \begin{minipage}{\textwidth}
  \begin{center}
      \includegraphics[width=0.53\textwidth]{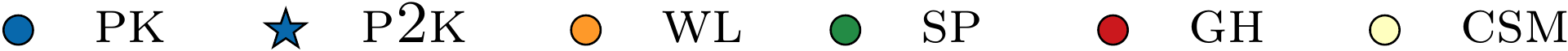}\hspace{0.4cm}
      \includegraphics[width=0.41\textwidth]{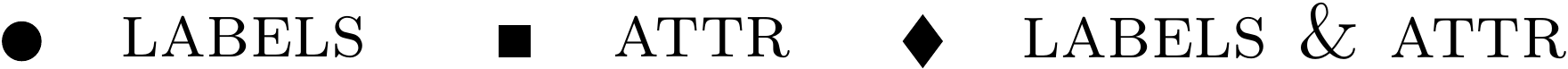}
  \end{center}
  \end{minipage}
  \begin{minipage}{\textwidth}
  \subfloat[\textsc{synthetic}]{\includegraphics[width=0.495\textwidth]{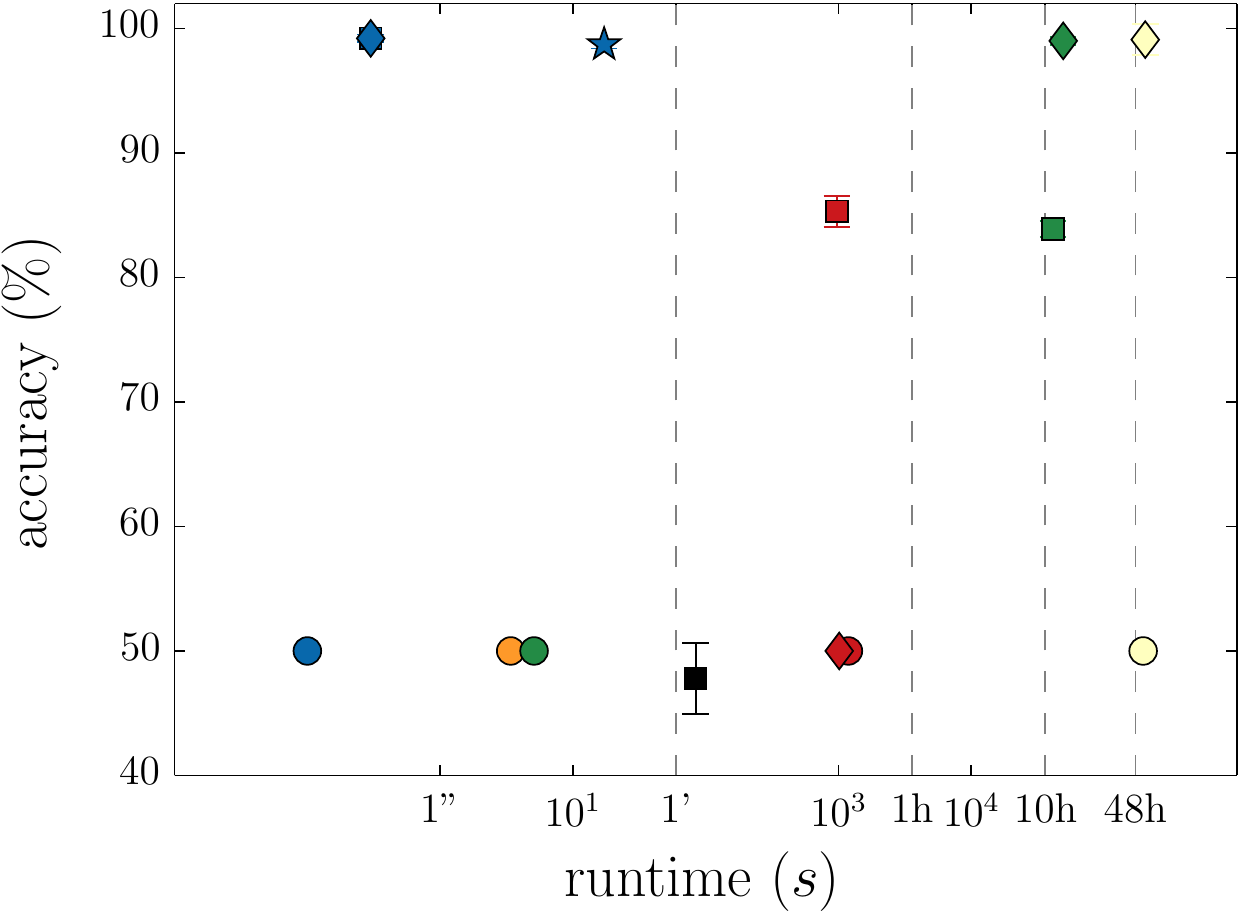}}
  \hfill
  \subfloat[\textsc{enzymes}]{\includegraphics[width=0.49\textwidth]{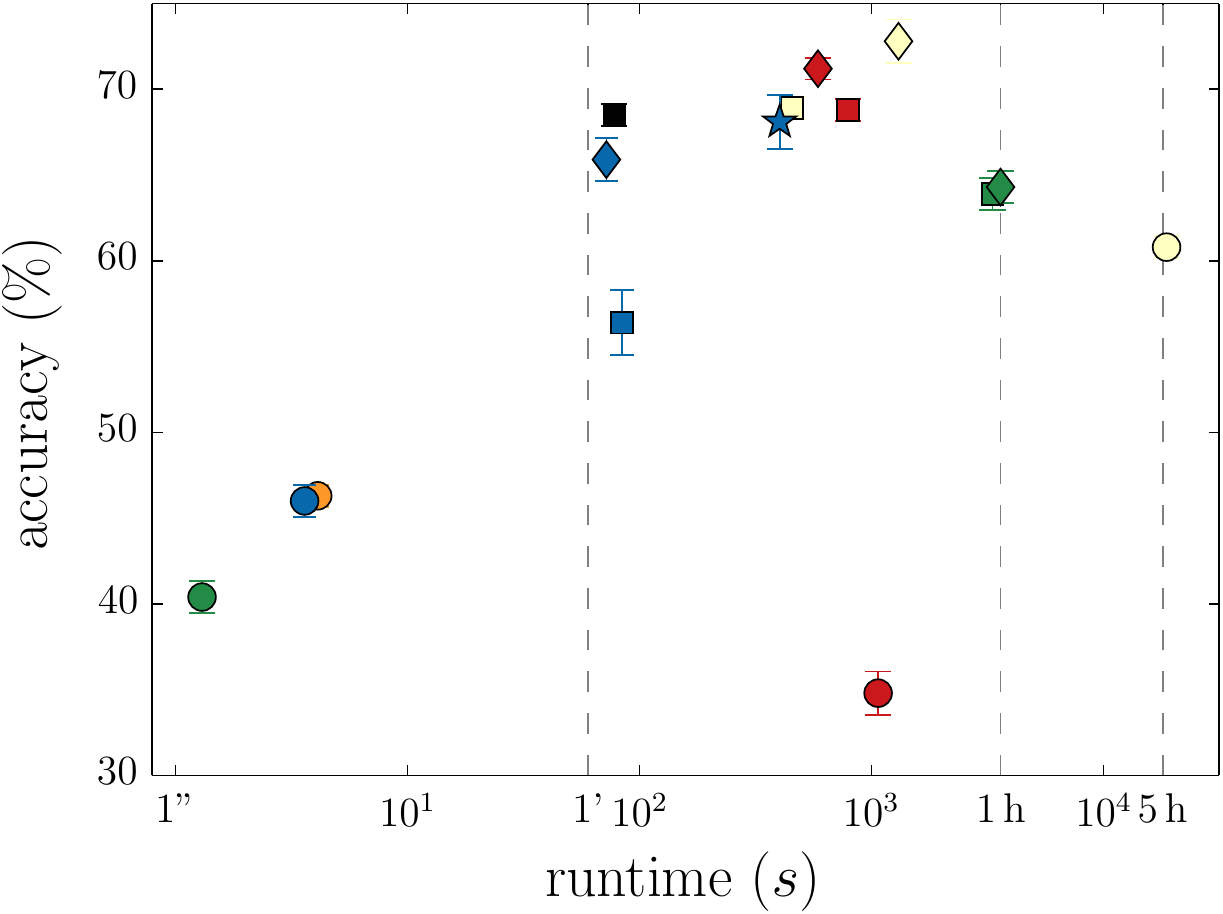}} \\
  \subfloat[\textsc{bzr}]{\includegraphics[width=0.49\textwidth]{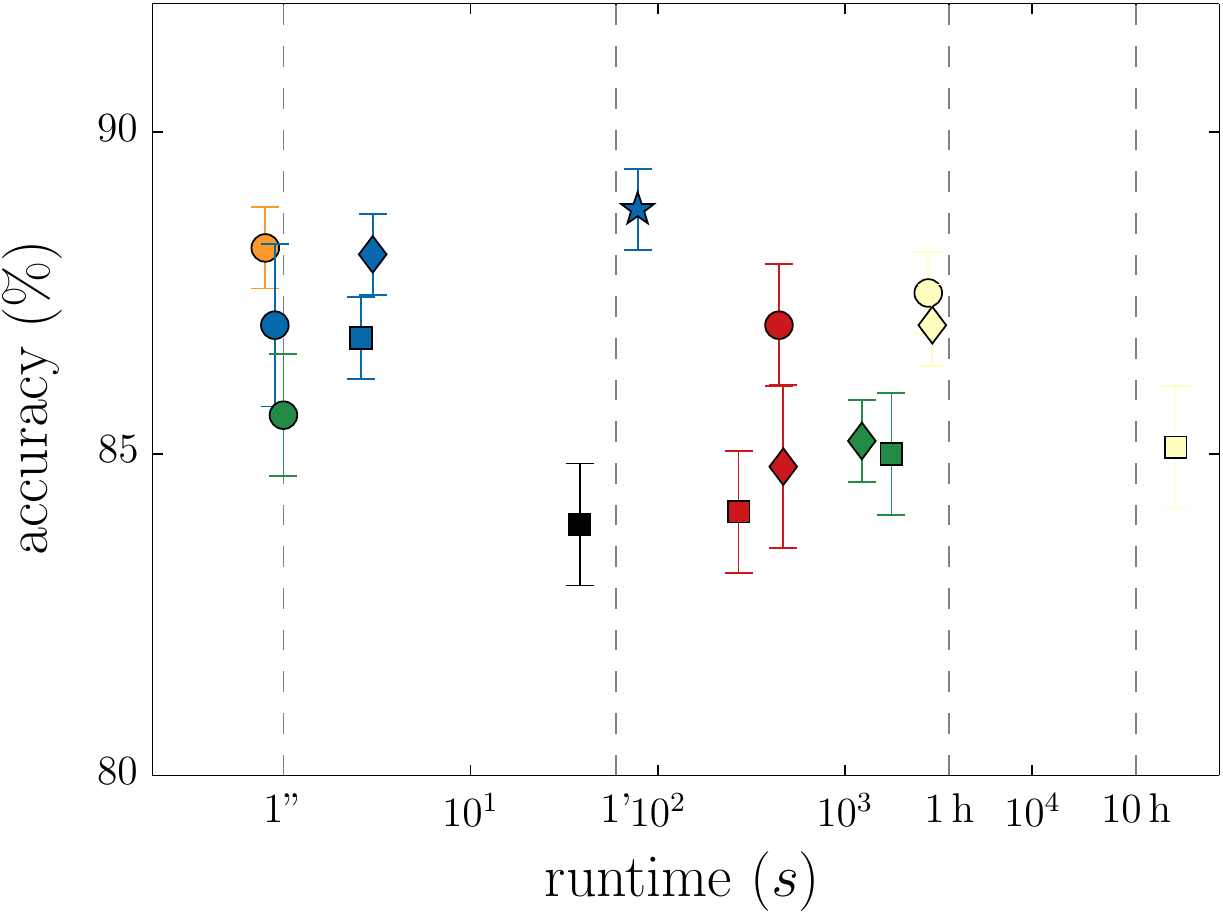}}
  \hfill
  \subfloat[\textsc{pro-full}]{\includegraphics[width=0.49\textwidth]{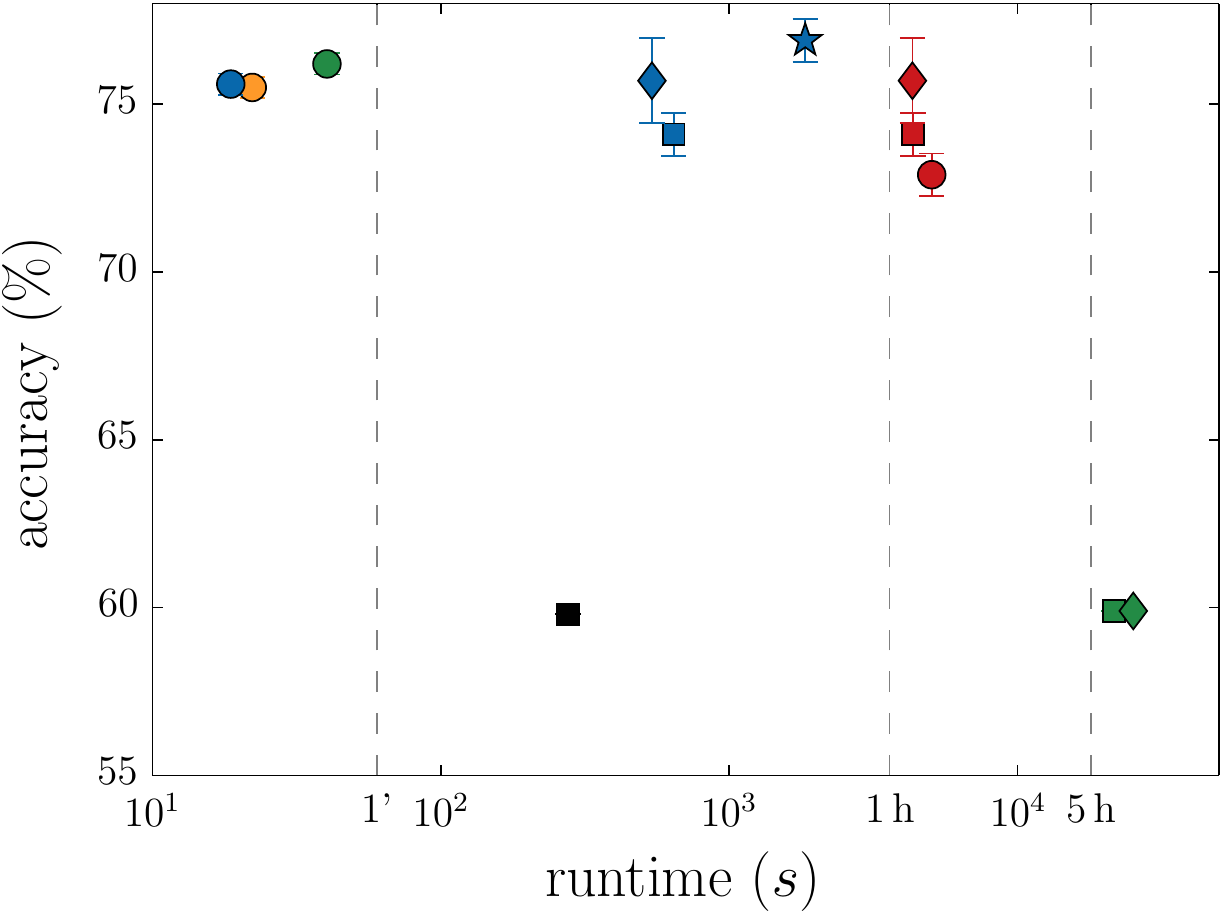}}
  \end{minipage}
  \caption[]{\textbf{Attributed Graphs -- Log Runtime vs.\ Accuracy.}
    The plots show log(runtimes) in seconds plotted against average
    accuracy ($\pm$ standard deviation). Methods are encoded by color
    and the used information (labels, attributes or both) is encoded
    by shape. Note that \textsc{p2k} also uses both, labels and
    attributes, however in contrast to \textsc{pk} both are
    propagated. For \textsc{pro-full} the \textsc{csm} kernel
    computation exceeded 32GB of memory.  On \textsc{synthetic}
    \textsc{csm} using the attribute information only could not be
    computed within $72$h.}
  \label{fig:timeVSacc}
  \end{center}
\end{figure}

\paragraph{Attributed Graphs\\}
The experimental results for various datasets with attributed graphs
are illustrated in Figure~\ref{fig:timeVSacc}.  The plots show runtime
versus average accuracy, where the error bars reflect standard
deviation of the accuracies.  As we are interested in good predictive
performance while achieving fast kernel computation, methods in the
upper-left corners provide the best performance with respect to both
quality and speed.  For \textsc{pk}, \textsc{sp}, \textsc{gh}, and
\textsc{csm} we compare three variants: one where we use the labels
only, one where we use the attribute information only, and one where
both labels and attributes are used. \textsc{wl} is computed with
label information only. For \textsc{synthetic},
cf.\ Figure~\ref{fig:timeVSacc}(a), we used the node degree as label
information. Further, we compare the performance of \textsc{p2k},
which propagates labels and attributes as described in
Section~\ref{sec:attributes}.  Detailed results on \textsc{synthetic}
and all bioinformatics datasets are provided in
Table~\ref{tab:acc_attr} (average accuracies) and
Table~\ref{tab:time_attr} (runtimes) in the Appendix.  From
Figure~\ref{fig:timeVSacc} we clearly see that propagation kernels
tend to appear in the upper-left corner, that is, they are achieving
good predictive performance while being fast, leading to a positive
answer of question {\bf (Q4)}.  Note that the runtimes are shown on a
log scale.  We can also see that \textsc{p2k}, propagating both labels
and attributes, (blue star) usually outperforms the the simple
\textsc{pk} implementation not considering attribute arrangements
(blue diamond). However, this comes at the cost of being slower. So,
we can use the flexibility of propagation kernels to trade predictive
quality against speed or vice versa according to the requirements of
the application at hand. This supports a positive answer to question
{\bf (Q3)}.

\subsubsection{Graph Classification on Novel Applications}
The flexibility of propagation kernels arising from easily
interchangeable propagation schemes and their efficient computation
via \textsc{lsh} allows us to apply graph kernels to novel domains.
First, we are able to compare larger graphs with reasonable time
expended, opening up the use of graph kernels for object category
prediction of 3\textsc{d} point clouds in the context of robotic
grasping \citep{Neumann13mlg}.  Depending on their size and the
perception distance, point clouds of household objects can easily
consist of several thousands of nodes. Traditional graph kernels
suffer from enormous computation times or memory problems even on
datasets like \textsc{db}, which can still be regarded medium sized.
These issues aggravate even more when considering image data. So far,
graph kernels have been used for image classification on the scene
level where the nodes comprise segments of similar pixels and one
image is then represented by less than 100 so-called
superpixels. Utilizing off-the-shelf techniques for efficient
diffusion on grid graphs allows the use of propagation kernels to
analyze images on the pixel level and thus opens up a whole area of
interesting problems in the intersection of graph-based machine
learning and computer vision.  As a first step, we apply graph
kernels, more precisely propagation kernels, to texture
classification, where we consider datasets with thousands of graphs
containing a total of several millions of nodes.

\paragraph{3\textsc{d} Object Category Prediction\\}
In this set of experiments, we follow the protocol introduced in
\citep{Neumann13mlg}, where the graph kernel values are used to derive
a prior distribution on the object category for a given query object.
The experimental results for the 3\textsc{d}-object classification are
summarized in Table~\ref{tab:acc_obj}.  We observe that propagation
kernels easily deal with the point-cloud graphs. From the set of
baseline graph kernels considered, only \textsc{wl} was feasible to
compute, however with poor performance. Propagation kernels clearly
benefit form their flexibility as we can improve the classification
accuracy from $75.4\%$ to $80.7\%$ when considering the object
curvature attribute. These results are extremely promising given that
we tackle a classification problem with 11 classes having only 40
training examples for each query object.

\begin{table}[!t]
  \centering
  \caption{\textbf{Point Cloud Graphs.} Average accuracies of
    \textsc{loo} cross validation on \textsc{db}. The reported
    standard errors refer to 10 kernel recomputations.  Runtimes are
    given in sec ($x''$).  The kernel parameters, $t_{\textsc{max}}$
    for all \textsc{pk}s and $h_{\textsc{max}}$ for \textsc{wl}, were
    learned on the training splits ($t_{\textsc{max}},
    h_{\textsc{max}} \in \{0, \dots 15\}$).  \textsc{a lsh}
    corresponds to \textsc{pk} with $t_{\textsc{max}}=0$.
    \textsc{sp}, \textsc{gh}, and \textsc{csm} were either \scriptsize
    OUT OF MEMORY \normalsize or the computation did not finish within
    $24$h for all settings.}
  \vspace*{1ex}
  \begin{tabular}{ccccccc}
    \toprule
    &\multicolumn{2}{c}{\textsc{labels}}	& \multicolumn{2}{c}{\textsc{labels}}  \\
    &&					& \multicolumn{2}{c}{\textsc{attr}} & \multicolumn{2}{c}{\textsc{attr}}  \\
    \cmidrule(l){2-7}
    &\textsc{pk}	& \textsc{wl}	     	& \textsc{pk}		& \textsc{p2k}			& \textsc{a}		& \textsc{a lsh}\\
    \midrule
    acc$ \pm $stderr	&$75.6 \pm 0.6$	&$70.7 \pm 0.0$		&$76.8 \pm 1.3$		&$\mathbf{82.9} \pm 0.0$	&$36.4 \pm 0.0$ 	&$63.4 \pm 0.0$\\
    runtime			& $0.2''$ 	&$0.4''$	 	&$0.3''$ 		&$34.8''$ 			&$40.0''$		&$0.0''$\\
    \bottomrule
  \end{tabular}
  \label{tab:acc_obj}
\end{table}

\paragraph{Grid Graphs\\}
For \textsc{brodatz} and \textsc{plants} we follow the experimental
protocol in \citep{NeumannHKKB14}.  The \textsc{pk} parameter
$t_{\textsc{max}}$ was learned on a training subset of the full
dataset ($t_{\textsc{max}} \in \{0,3,5,8,10,15,20\}$).  For
\textsc{plants} this training dataset consists of 10\% of the full
data; for \textsc{brodatz} we used 20\% of the \textsc{brodatz-o-r}
data as the number of classes in this dataset is much larger (32
textures).  We also learned the quantization values ($col \in
\{3,5,8,10,15\}$) and neighborhoods ($B \in \{N_{1,4}$, $N_{1,8}$,
$N_{2,16}\}$, cf.\ Eq.~\eqref{equ:circ_neigh}).  For \textsc{brodatz}
the best performance on the training data was achieved with $3$ colors
and a 8-neighborhood, whereas for \textsc{plants} $5$ colors and the
4-neighborhood was learned.  We compare \textsc{pk} to the simple
baseline labels using label counts only and to a powerful second-order
statistical feature based on the gray-level co-occurrence matrix
\citep{haralick73} comparing intensities (\textsc{glcm-gray}) or
quantized labels (\textsc{glcm-quant}) of neighboring pixels.

The experimental results for grid graphs
are shown in Table~\ref{tab:acc_texture}.  While not outperforming
sophisticated and complex state-of-the-art computer vision approaches
to texture classification, we find that it is feasible to compute
\textsc{pk}s on huge image datasets achieving respectable performance
out of the box. This is -- compared to the immense tuning of features
and methods commonly done in computer vision -- a great success.  On
\textsc{plants}, \textsc{pk} achieves an average accuracy of $82.5\%$,
where the best reported result so far is $83.7\%$, which was only
achieved after tailoring a complex feature
ensemble~\citep{NeumannHKKB14}.  In conclusion, propagation kernels are
an extremely promising approach in the intersection of machine
learning, graph mining, and computer vision.

\begin{table}[!t]
  \centering
  \caption{\textbf{Grid Graphs.} Average accuracies $\pm$ standard
    errors of 10-fold \textsc{cv} (10 runs).  The \textsc{pk}
    parameter $t_{\textsc{max}}$ as well as color quantization and
    pixel neighborhood was learned on a training subset of the full
    dataset.  Average runtimes in sec ($x''$) or min ($x'$) given in
    parentheses refer to the learned parameter settings. For
    \textsc{glcm-quant} and \textsc{labels} the same color
    quantization as for \textsc{pk} was applied.  \textsc{labels}
    corresponds to \textsc{pk} with $t_{\textsc{max}}=0$.
    \textsc{brodatz-o-r} is using the original images and their 90
    degree rotated versions and \textsc{brodatz-o-r-s-rs} additionally
    includes their scaled, and scaled and rotated versions.}
  \vspace*{1ex}
  \begin{tabular}{p{2.2cm}ccc}
    \toprule
    & \multicolumn{3}{c}{dataset} \\
    \cmidrule(l){2-4}
    method 	 		&\textsc{brodatz-o-r} 			& \textsc{brodatz-o-r-s-rs}  		& \textsc{plants}	\\
    \midrule
    \textsc{pk}   		&$\mathbf{89.6}$ $\pm 0.0$ ($3.5'$) 	& $\mathbf{85.7}$ $\pm 0.0$ ($7.1'$) 	&$\mathbf{82.5}$ $\pm 0.1$ ($3.0'$)\\
    \textsc{labels}  	&$5.0 $ $\pm 0.0$ ($1.1'$)		& $4.9$ $\pm 0.0$  ($2.2'$) 		&$59.5$ $\pm 0.0$ ($11.5''$) \\
    \midrule
    \textsc{glcm-gray}  	& $87.2$ $\pm 0.0$ ($29.5''$) 		& $79.4$ $\pm 0.0$ ($44.8''$)		& $76.6$ $\pm 0.0$ ($1.4'$)\\
    \textsc{glcm-quant} 	& $78.6$ $\pm 0.0$ ($24.9''$)  	& $68.6$ $\pm 0.0$ ($44.8''$)  	& $37.5$ $\pm 0.0$ ($1.1'$)\\
    \bottomrule
  \end{tabular}
  \label{tab:acc_texture}
\end{table}

Summarizing all experimental results, the capabilities claimed in
Table~\ref{tab:kernels} are supported.  Propagation kernels have
proven extremely flexible and efficient and thus question {\bf (Q3)}
can ultimately be answered affirmatively.

\section{Conclusion}
Random walk-based models provide a principled way of spreading
information and even handling missing and uncertain information within
graphs. Known labels are, for example, propagated through the graph in
order to label all unlabeled nodes.  In this paper, we showed how to
use random walks to discover structural similarities shared between
graphs for the construction of a graph kernel, namely the propagation
kernel.  Intuitively, propagation kernels count common
sub-distributions induced in each iteration of running inference in
two graphs leading to the insight that graph kernels are much closer
to graph-based learning than assumed before.

As our experimental results demonstrate, propagation kernels are
competitive in terms of accuracy with state-of-the-art kernels on
several classification benchmark datasets of labeled and attributed
graphs. In terms of runtime, propagation kernels outperform all
recently developed efficient and scalable graph kernels. Moreover,
being tied to the propagation scheme, propagation kernels can be
easily adapted to novel applications not having been tractable for
graph kernels before.

Propagation kernels provide several interesting avenues for future
work.  While we have used classification to guide the development of
propagation kernels, the results are directly applicable to
regression, clustering, and ranking, among other tasks. Employing
message-based probabilistic inference schemes such as (loopy) belief
propagation directly paves the way to dealing with graphical models.
Exploiting that graph kernels and graph-based learning (learning on
the node level) are closely related, a natural extension to this work
is the derivation of a unifying propagation-based framework for
structure representation independent of the learning task being on the
graph or node level.

\bibliography{propBib}{}

\pagebreak

\appendix

\section{Runtimes for Partially Labeled Graphs}

\begin{figure}[!h]
  \begin{center}
    \includegraphics[width=0.6\textwidth]{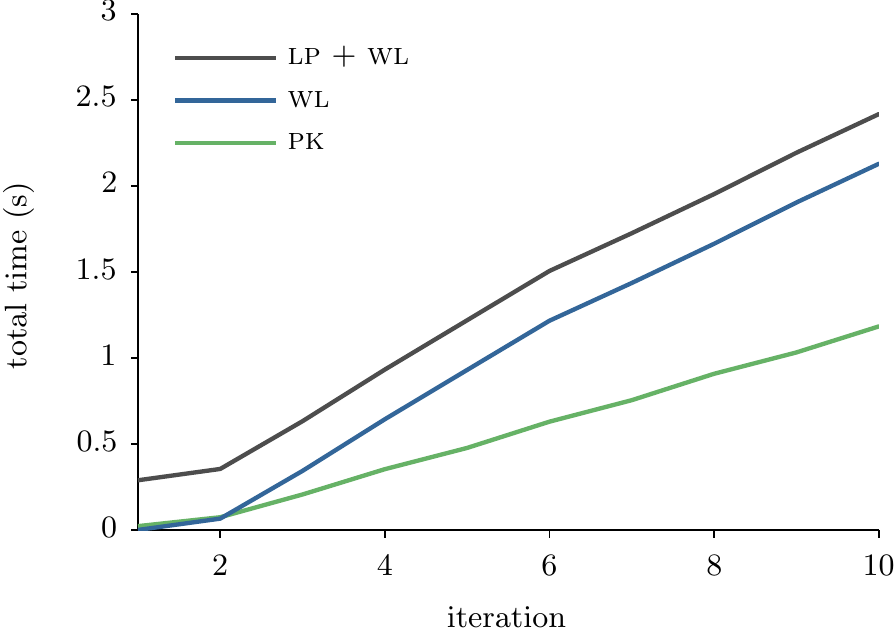}
      \vspace*{-1ex}
    \caption[]{\textbf{Runtime for Partially Labeled \textsc{msrc21}.}
      Avg. time in secs over $10$ instances of the \textsc{msrc21}
      dataset with $50\%$ labeled nodes for kernel iterations from $0$
      to $10$. We compare \textsc{pk} to \textsc{wl} with unlabeled
      nodes treated as additional label and with hard labels derived
      from converged label propagation (\textsc{lp}$+$\textsc{wl}).  }
    \label{fig:image_run}
  \end{center}
\end{figure}

\section{Detailed Results on Attributed Graphs}

\begin{table}[!h]
  \centering
  \caption{\textbf{Attributed Graphs -- Runtimes.} Kernel computation
    times (cputime) are given in sec ($x''$), min ($x'$), or hours
    ($x$h).  For all \textsc{pk}s, $t_{\textsc{max}} = 10$; for
    \textsc{wl}, $h_{\textsc{max}} = 10$; and for \textsc{csm}, $k=7$.
    All computations are performed on machines with 3.4\,GHz Intel
    core i7 processors.  Note that \textsc{csm} is implemented in
    Java, so comparing computation times is only possible to a limited
    extent.  \scriptsize OUT OF TIME \normalsize means that the
    computation did not finish within $72$h.}
  \vspace*{-1ex}
  \begin{tabular}{p{0.01cm}p{0.01cm}p{0.3cm}cccccccc}
    \toprule
    &&& \multicolumn{7}{c}{dataset} \\
    \cmidrule(l){4-10}
    \multicolumn{3}{c}{method} 	 &\textsc{synthetic} 	& \textsc{enzymes} 		& \textsc{proteins} & \textsc{pro\_full} &\textsc{bzr} 		& \textsc{cox}{\footnotesize 2} & \textsc{dhfr} 	\\
    \midrule
    \multirow{5}{*}{\begin{turn}{90}\textsc{labels}\end{turn}} &
    &\textsc{pk}	&$\mathbf{0.1''}$&$\mathbf{3.6''}$&$\mathbf{17.8''}$&$\mathbf{18.7''}$&$0.9''$&$1.1''$&$3.4''$\\
    &&\textsc{wl}	&$3.4''$&$4.1''$&$25.9''$&$22.2''$&$\mathbf{0.8''}$&$1.0''$&$4.2''$\\
    &&\textsc{sp}	&$5.1''$&$1.3''$&$38.7''$&$40.3''$&$1.0''$&$1.2''$&$\mathbf{2.3''}$\\
    &&\textsc{gh}	&$19.8'$&$17.8'$&$2.2$h  &$1.4$h  &$7.4'$ &$10.8'$&$29.5'$\\
    &&\textsc{csm}	&$54.8$h&$5.2$h & -- & -- & $46.4'$ & $1.6$h & $3.7$h  \\
    \midrule
    &\multirow{4}{*}{\begin{turn}{90}\textsc{attr}\end{turn}} &
    \textsc{pk} 		&$\mathbf{0.3''}$&$\mathbf{1.4'}$&$\mathbf{23.6''}$&$\mathbf{10.7'}$&$\mathbf{2.6''}$&$\mathbf{2.8''}$&$\mathbf{14.8''}$\\	
    &&\textsc{sp}   &$11.5$h&$55.4'$&$5.9$h&$6.0$h&$29.5'$&$25.9'$&$1.3$h\\
    &&\textsc{gh} 	&$16.2'$&$13.2'$&$1.9$h&$72.4'$&$4.5'$&$6.6'$&$15.8'$\\
    &&\textsc{csm} & \tiny OUT OF TIME& $7.6'$ & -- & -- & $16.2$h & $31.5$h & $97.5$h  \\
    \midrule
    \multirow{5}{*}{\begin{turn}{90}\textsc{labels}\end{turn}} &
    \multirow{5}{*}{\begin{turn}{90}\textsc{attr}\end{turn}} &
    \textsc{pk} 	&$\mathbf{0.3''}$&$\mathbf{1.2'}$&$\mathbf{20.0''}$&$\mathbf{9.0'}$&$\mathbf{3.0''}$&$\mathbf{3.5''}$&$\mathbf{15.1''}$\\
    &&\textsc{p2k}	&$17.2''$&$6.7'$ &$31.4'$&$30.6'$&$1.3'$&$1.8'$&$5.9'$\\
    &&\textsc{sp}	&$13.7$h &$1.0$h &$6.8$h &$7.0$h&$20.5'$&$32.9'$&$1.6$h\\
    &&\textsc{gh}	&$16.9'$ &$9.8'$ &$1.2$h &$1.2$h&$7.8'$&$11.7'$&$31.2'$\\
    &&\textsc{csm}	&$56.8$h &$21.8'$& --    & -- & $48.8'$ & $1.6$h & $3.7$h  \\
    \midrule
    &\multirow{2}{*}{\begin{turn}{90}\textsc{attr}\end{turn}} &
    \multirow{2}{*}{\textsc{a}} &\multirow{2}{*}{$1.4'$}&\multirow{2}{*}{$1.3'$}&\multirow{2}{*}{$6.4'$}&\multirow{2}{*}{$4.6'$}&\multirow{2}{*}{$38.3''$}&\multirow{2}{*}{$1.1'$}&\multirow{2}{*}{$3.2'$}\\
    &&  \\
    \bottomrule
  \end{tabular}
  \label{tab:time_attr}
\end{table}

\begin{landscape}
  \begin{table}[!ht]
    \centering
    \caption{\textbf{Attributed Graphs -- Accuracies.} Average
      accuracies $\pm$ standard error of 10-fold cross validation (10
      runs).  All \textsc{pk}s were also recomputed for each run as
      there is randomization in the \textsc{lsh} computation. The
      kernel parameters $t_{\textsc{max}}$ for all \textsc{pk}s and
      $h_{\textsc{max}}$ for \textsc{wl} were learned on the training
      splits ($t_{\textsc{max}}, h_{\textsc{max}} \in \{0,1, \dots
      10\}$).  For all \textsc{pk}s we applied standardization to the
      attributes and one hash per attribute dimension is computed.
      Whenever the normalized version of a kernel performed better
      than the unnormalized version we report these results and mark
      the method with $\ast$.  \textsc{csm} is implemented in Java and
      computations were performed on a machine with 32\acro{GB} of
      memory.  \scriptsize OUT OF MEMORY \normalsize indicates a Java
      \texttt{outOfMemeoryError}.  Bold indicates that the method
      performs significantly better than the second best method under
      a paired $t$-test ($p<0.05$). The \textsc{svm} cost parameter is
      learned on the training splits. We choose $c \in \{10^{-7},
      10^{-5}, \cdots 10^{5},10^{7} \}$ for normalized kernels and $c
      \in \{10^{-7}, 10^{-5}, 10^{-3},10^{-1} \}$ for unnormalized
      kernels.}
    \vspace*{1ex}
    \begin{tabular}{p{0.01cm}p{0.01cm}p{0.3cm}cccccccc}
      \toprule
      &&& \multicolumn{7}{c}{dataset} \\
      \cmidrule(l){4-10}
      \multicolumn{3}{c}{method} 	 &\textsc{synthetic} 	& \textsc{enzymes} & \textsc{proteins} & \textsc{pro-full} &\textsc{bzr}	& \textsc{cox}{\footnotesize 2} & \textsc{dhfr} 	\\
      \midrule

      \multirow{5}{*}{\begin{turn}{90}\textsc{labels}\end{turn}} &
      &\textsc{pk}  	&$50.0$ $\pm 0.0^\ast$&$46.0$ $\pm 0.3$&$75.6$ $\pm 0.1^\ast$&$75.6$ $\pm 0.1^\ast$&$87.0$ $\pm 0.4$&$81.0$ $\pm 0.2$&$83.5$ $\pm 0.2^\ast$\\
      &&\textsc{wl} 	&$50.0$ $\pm 0.0^\ast$&$46.3$ $\pm 0.2^\ast$&$75.5$ $\pm 0.1^\ast$&$75.5$ $\pm 0.1^\ast$&$88.2$ $\pm 0.2$&$\mathbf{83.2}$ $\pm 0.2^\ast$&$84.1$ $\pm 0.2$\\
      &&\textsc{sp}  	&$50.0$ $\pm 0.0^\ast$&$40.4$ $\pm 0.3^\ast$&$76.2$ $\pm 0.1^\ast$&$76.2$ $\pm 0.1^\ast$&$85.6$ $\pm 0.3$&$81.0$ $\pm 0.4^\ast$&$82.0$ $\pm 0.3^\ast$\\
      &&\textsc{gh}	&$50.0$ $\pm 0.0^\ast$&$34.8$ $\pm 0.4^\ast$&$72.9$ $\pm 0.2^\ast$&$72.9$ $\pm 0.2^\ast$&$87.0$ $\pm 0.3$&$81.4$ $\pm 0.3^\ast$&$79.5$ $\pm 0.4$\\
      &&\textsc{csm}	&$50.0$ $\pm 0.0^\ast$&$60.83$ $\pm 0.2$&\tiny OUT OF MEMORY&\tiny OUT OF MEMORY&$87.5$ $\pm 0.2^\ast$&$80.7$ $\pm 0.3$&$82.6$ $\pm 0.2^\ast$\\
      \midrule
      &\multirow{4}{*}{\begin{turn}{90}\textsc{attr}\end{turn}} &
      \textsc{pk}	&$99.2$ $\pm 0.1^\ast$&$56.4$ $\pm 0.6$&$72.7$ $\pm 0.2$&$74.1$ $\pm 0.2^\ast$&$86.8$ $\pm 0.2^\ast$&$78.2$ $\pm 0.4$&$84.3$ $\pm 0.1$\\
      &&\textsc{sp}	&$83.9$ $\pm 0.2^\ast$&$63.9$ $\pm 0.3^\ast$&$74.3$ $\pm 0.1^\ast$&$59.9$ $\pm 0.0^\ast$&$85.0$ $\pm 0.3^\ast$&$78.2$ $\pm 0.0$&$78.9$ $\pm 0.3^\ast$\\
      &&\textsc{gh}	&$85.3$ $\pm 0.4$&$68.8$ $\pm 0.2^\ast$&$72.6$ $\pm 0.1^\ast$&$61.2$ $\pm 0.0^\ast$&$84.1$ $\pm 0.3^\ast$&$79.5$ $\pm 0.2^\ast$&$79.0$ $\pm 0.2$\\
      &&\textsc{csm}	& -- & $68.9$ $\pm 0.2^\ast$&\tiny OUT OF MEMORY&\tiny OUT OF MEMORY&$85.1$ $\pm 0.3^\ast$&$77.6$ $\pm 0.3$&$79.5$ $\pm 0.2$\\
      \midrule
      \multirow{5}{*}{\begin{turn}{90}\textsc{labels}\end{turn}} &
      \multirow{5}{*}{\begin{turn}{90}\textsc{attr}\end{turn}} &
      \textsc{pk}  	&$99.2$ $\pm 0.1^\ast$&$65.9$ $\pm 0.4^\ast$&$76.3$ $\pm 0.2^\ast$&$75.7$ $\pm 0.4^\ast$&$88.1$ $\pm 0.2^\ast$&$79.4$ $\pm 0.6$&$84.1$ $\pm 0.3^\ast$\\
      &&\textsc{p2k}  &$98.7$ $\pm 0.1$&$68.1$ $\pm 0.5$&$75.9$ $\pm 0.2^\ast$&$\mathbf{76.9}$ $\pm 0.2^\ast$&$88.8$ $\pm 0.2$&$80.9$ $\pm 0.4$&$83.5$ $\pm 0.3^\ast$\\
      &&\textsc{sp} 	&$99.0$ $\pm 0.1$&$64.3$ $\pm 0.3^\ast$&$73.2$ $\pm 0.2^\ast$&$59.9$ $\pm 0.0^\ast$&$85.2$ $\pm 0.2^\ast$&$78.5$ $\pm 0.1$&$79.7$ $\pm 0.2^\ast$\\
      &&\textsc{gh} 	&$50.0$ $\pm 0.0^\ast$&$71.2$ $\pm 0.2^\ast$&$73.0$ $\pm 0.1^\ast$&$60.9$ $\pm 0.0^\ast$&$84.8$ $\pm 0.4$&$79.5$ $\pm 0.2^\ast$&$80.0$ $\pm 0.2$\\
      &&\textsc{csm}	&$99.0$ $\pm 0.1$&$\mathbf{72.8}$ $\pm 0.4^\ast$&\tiny OUT OF MEMORY&\tiny OUT OF MEMORY&$87.0$ $\pm 0.2^\ast$&$79.2$ $\pm 0.4^\ast$&$80.1$ $\pm 0.3^\ast$\\
      \midrule
      &\multirow{2}{*}{\begin{turn}{90}\textsc{attr}\end{turn}} &
      \multirow{2}{*}{\textsc{a}} 	&\multirow{2}{*}{$47.8$ $\pm 0.9^\ast$}&\multirow{2}{*}{$68.5$ $\pm 0.2^\ast$}&\multirow{2}{*}{$72.8$ $\pm 0.2^\ast$}&\multirow{2}{*}{$59.8$ $\pm 0.0^\ast$}&\multirow{2}{*}{$83.9$ $\pm 0.3^\ast$}&\multirow{2}{*}{$78.2$ $\pm 0.0$}&\multirow{2}{*}{$74.8$ $\pm 0.2^\ast$}\\
      &\\
      \bottomrule
    \end{tabular}
    \label{tab:acc_attr}
  \end{table}
\end{landscape}

\end{document}